\def\year{2021}\relax
\documentclass[letterpaper]{article} % DO NOT CHANGE THIS
\usepackage{aaai21}  % DO NOT CHANGE THIS
\usepackage{times}  % DO NOT CHANGE THIS
\usepackage{helvet} % DO NOT CHANGE THIS
\usepackage{courier}  % DO NOT CHANGE THIS
\usepackage[hyphens]{url}  % DO NOT CHANGE THIS
\usepackage{graphicx} % DO NOT CHANGE THIS
\usepackage{natbib}  % DO NOT CHANGE THIS AND DO NOT ADD ANY OPTIONS TO IT
\usepackage{caption} % DO NOT CHANGE THIS AND DO NOT ADD ANY OPTIONS TO IT
\usepackage{color,soul}

\usepackage[normalem]{ulem}
\title{Transfer Learning for Efficient Iterative Safety Validation}
\author {
        Anthony Corso\textsuperscript{\rm 1} and
        Mykel J. Kochenderfer\textsuperscript{\rm 1}\\
}
\affiliations {
        \textsuperscript{\rm 1}Stanford University, Department of Aeronautics and Astronautics, 496 Lomita Mall, Stanford, CA 94305 \\
        \{acorso, mykel\}@stanford.edu
}
\usepackage{todonotes}
\usepackage{tikz}
\usetikzlibrary{arrows,fit} % for problem diagram
\usetikzlibrary{positioning} % for problem diagram
\usetikzlibrary{arrows.meta,calc,shapes}

\setul{0.8ex}{0.4ex}

\usepackage{pgfplots}
\pgfplotsset{compat=newest}
\usepgfplotslibrary{groupplots}
\usepgfplotslibrary{polar}
\usepgfplotslibrary{smithchart}
\usepgfplotslibrary{statistics}
\usepgfplotslibrary{dateplot}
\usetikzlibrary{arrows.meta}
\usetikzlibrary{backgrounds}
\usepgfplotslibrary{patchplots}
\usepgfplotslibrary{fillbetween}
\pgfplotsset{%
layers/standard/.define layer set={%
    background,axis background,axis grid,axis ticks,axis lines,axis tick labels,pre main,main,axis descriptions,axis foreground%
}{grid style= {/pgfplots/on layer=axis grid},%
    tick style= {/pgfplots/on layer=axis ticks},%
    axis line style= {/pgfplots/on layer=axis lines},%
    label style= {/pgfplots/on layer=axis descriptions},%
    legend style= {/pgfplots/on layer=axis descriptions},%
    title style= {/pgfplots/on layer=axis descriptions},%
    colorbar style= {/pgfplots/on layer=axis descriptions},%
    ticklabel style= {/pgfplots/on layer=axis tick labels},%
    axis background@ style={/pgfplots/on layer=axis background},%
    3d box foreground style={/pgfplots/on layer=axis foreground},%
    },
}

\pgfplotsset{
colormap={plots1}{rgb(0.00000000cm)=(1.00000000,0.00000000,0.00000000)
rgb(0.25000000cm)=(1.00000000,0.53200400,0.53200400)
rgb(0.50000000cm)=(1.00000000,1.00000000,1.00000000)
rgb(0.75000000cm)=(0.51697700,0.75967200,0.55607000)
rgb(1.00000000cm)=(0.00000000,0.50244900,0.08093390)},
}
\pgfplotsset{
colormap={plots1}{rgb(0.00000000cm)=(1.00000000,0.00000000,0.00000000)
rgb(0.25000000cm)=(1.00000000,0.53200400,0.53200400)
rgb(0.50000000cm)=(1.00000000,1.00000000,1.00000000)
rgb(0.75000000cm)=(0.51697700,0.75967200,0.55607000)
rgb(1.00000000cm)=(0.00000000,0.50244900,0.08093390)},
}

\usepackage{amsmath}

\usepackage{amsfonts}
\usepackage{cleveref}
\usepackage{dsfont}
\usepackage{gensymb}
\usepackage{siunitx}
\usepackage{booktabs}
\usepackage{subcaption}
\usepackage[switch]{lineno}

\makeatletter
\usepackage{etoolbox}
\patchcmd\H@refstepcounter{\protected@edef}{\protected@xdef}{}{}
\makeatother

\begin{document}
% \linenumbers  %
\maketitle

\begin{abstract}
Safety validation is important during the development of safety-critical autonomous systems but can require significant computational effort. Existing algorithms often start from scratch each time the system under test changes. We apply transfer learning to improve the efficiency of reinforcement learning based safety validation algorithms when applied to related systems. Knowledge from previous safety validation tasks is encoded through the action value function and transferred to future tasks with a learned set of attention weights. Including a learned state and action value transformation for each source task can improve performance even when systems have substantially different failure modes. We conduct experiments on safety validation tasks in gridworld and autonomous driving scenarios. We show that transfer learning can improve the initial and final performance of validation algorithms and reduce the number of training steps.

\end{abstract}

\section{Introduction}
% Intro the problem of safety validation
Introducing autonomy into safety-critical domains, such as autonomous driving, aviation, and medicine, has the potential to improve both safety and efficiency. The consequences of operational errors of these systems include loss of property or human life, so extensive safety validation and testing is required before deployment. Black-box sampling approaches have emerged as a scalable safety validation tool for discovering failures in complex environments. Many algorithms for safety validation have been explored in the literature~\cite{corso2020survey}, often with the goal of finding failures of a system with fewer samples or less computational effort. There has been little focus on the potential efficiency of validating many related systems sequentially.

Designing and certifying safety-critical systems generally involves assessing a sequence of closely related systems. Safety validation of complex systems often requires significant computational effort and existing approaches generally start from scratch each time the system under test is changed. The need to frequently perform safety validation on related systems therefore imposes a large computational burden, but also an opportunity to improve safety validation efficiency. 

% What we did
To improve the efficiency of safety validation across related systems, we use knowledge from the validation of previous systems to inform the validation of the next system. We formulate iterative safety validation as a transfer learning problem by modeling each safety validation task as a Markov decision process. The previously solved tasks are used as the set of source tasks, and we transfer knowledge to future tasks in the form of action value functions. We use state-dependent attention weights to learn which previous solutions are applicable to the current problem.

Existing safety validation algorithms use approaches from optimization~\cite{Mathesen2019falsification}, path-planning~\cite{zutshi2014multiple}, reinforcement-learning~\cite{lee_adaptive_2020}, and importance sampling~\cite{huang2017accelerated}, and often only address the validation of a single system. \Citet{uesato2019rigorous} use previous versions of a system to train a failure classifier that predicts which initial conditions of a system will lead to failure, but their approach is not applicable to sequential decision making problems of the type we consider. \Citet{wang2020falsification} alternately train an agent and perform safety validation on it to improve robustness. On each iteration, the safety validation algorithm starts with the parameters from the previous iteration to improve efficiency. In fact, any parametric safety validation algorithm~\cite{koren2018adaptive, Akazaki2018falsification, kim2016improving} could simply reuse parameters from previous tasks and then fine-tune them for better performance. We demonstrate in our experiments, however, that a fine-tuning approach often fails to reach the same performance as starting from scratch.

To investigate the effectiveness of transfer learning in the safety validation setting, we apply existing transfer learning algorithms to the problem. These include fine-tuning of past solutions, and the attend, adapt, and transfer (A2T) algorithm~\cite{rajendran2017attend}. When the systems we wish to validate have dissimilar behavior, however, we find that existing approaches can perform poorly. We propose a modification to A2T that transforms the state and action value spaces for each source task to increase knowledge transfer between dissimilar tasks. We evaluate the initial performance, the final performance, and the number of training steps required to reach the same performance as a no-transfer algorithm. We consider four iterative safety validation tasks in gridworld and autonomous driving scenarios and demonstrate that transfer learning has the potential to significantly improve the performance and sample efficiency of safety validation algorithms.

\section{Background}
In this section, we introduce Markov decision processes (MDPs), discuss knowledge sharing between related MDPs, and formulate safety validation as a sequence of MDPs.

\subsection{Markov Decision Processes}
% What is an MDP
A Markov decision process (MDP)~\cite{dmubook} is a model for sequential decision making problems defined by the tuple ($\mathcal{S}$, $\mathcal{A}$, $P$, $R$, $\gamma$). The state space $\mathcal{S}$ contains all possible states of the MDP and the action space $\mathcal{A}$ contains the possible actions of a decision making agent. At each step, the agent chooses an action and the MDP transitions to a new state $s^\prime$ with probability $P(s^\prime \mid s, a)$ and receives a reward $r = R(s, a, s')$ discounted by a factor $\gamma$ for each step. An agent's behavior is controlled by a policy $\pi$ that maps states to actions such that $a = \pi(s)$. The optimal policy $\pi^*$ maximizes the \emph{action value function} $Q^\pi(s,a)$, which is the expected sum of discounted rewards by taking action $a$ from state $s$, and then following policy $\pi$. 

% Introduce DQN as a solution approach
There are many approaches to solving for $\pi^*$~\cite{sutton2018reinforcement} but we focus on deep $Q$-learning (DQN)~\cite{mnih2015human}. In DQN, the optimal action value function is approximated by a deep neural network with parameters $\theta$, $Q(s,a; \theta) \approx Q^*(s,a)$. The $Q$-network tries to minimize the loss with respect to a target network with parameters $\theta^-$. The parameters are updated using gradient descent
\begin{equation}
    \theta \gets \theta - \alpha \nabla_{\theta} \mathbb{E}[L(y(s^\prime; \theta^-), Q(s,a;\theta)]
\end{equation}
where $\alpha$ is the learning rate and loss function $L$ can be the squared error or the Huber loss~\cite{huber1992robust}. The target is $y(s^\prime; \theta^-) = r + \gamma \underset{a^\prime}{\max} \  Q(s', a';  \theta^-)$ when $s^\prime$ is not terminal, and $y(s^\prime; \theta^-) = r$ when $s'$ is terminal. The target network parameters are periodically updated to $\theta$ after a specified number of training steps. During training, the expectation is computed from samples that are stored in an experience replay buffer that is prioritized by the temporal difference error of each sample~\cite{schaul2016prioritized}.

\subsection{Transfer Learning}
% What is transfer learning for RL?
Transfer learning is concerned with using knowledge gained by solving one task to improve the learning process in another related task~\cite{taylor2009transfer}. In reinforcement learning, each task is an MDP and each solution is a policy. When tasks have different state and action spaces, we require task mappings that relate the states and actions between tasks. Task mappings may be provided by a human~\cite{taylor2007transfer} or learned from data~\cite{taylor2008autonomous}. If the state and action spaces are the same, then tasks can share a variety of low-level information such as experience samples $(s, a, s', r)$, action value functions, policies, or models of the environment. High-level information, such as a set of options, shaping rewards or feature encodings, may also be transferred to improve learning on a new task. One challenge in transfer learning is \emph{negative transfer} where knowledge from one or more source tasks impairs the performance on the current task. Negative transfer can be mitigated using an attention mechanism, or a human oracle that decides which tasks are relevant~\cite{taylor2009transfer}. 

\begin{figure}
\centering
\input{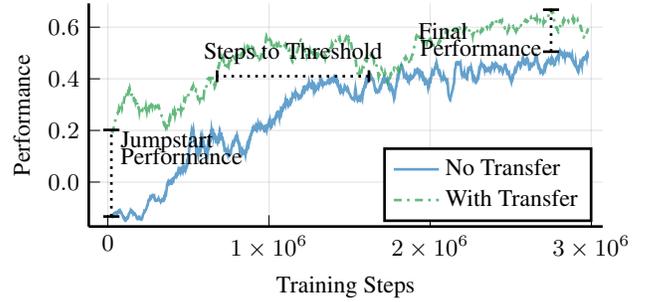}
\vspace{ -0.25in}
\caption{Metrics for evaluating transfer learning algorithms. }
\label{fig:transfer_metrics}
\end{figure}

% How are transfer learning algorithms evaluated
Transfer learning algorithms can be evaluated against a no-transfer alternative in a variety of ways (\cref{fig:transfer_metrics}). \emph{Jumpstart} is the amount of improvement before any training has occurred, \emph{final performance} is the difference between the best performances achieved, and \emph{steps to threshold} is the difference between the number of training steps required to reach a specified threshold.

\subsection{Safety Validation}
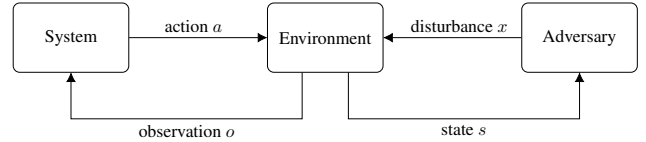
\begin{figure}
\centering
\resizebox{\columnwidth}{!}{% TikZ diagram for black-box safety validation problem formulation.

\tikzset{
    >={Latex[width=2mm,length=2mm]},
        base/.style = {rectangle, rounded corners, draw=black,
        minimum width=1cm, minimum height=1cm,
        text centered},
    block/.style = {base, minimum width=2.5cm, minimum height=1.5cm},
    sutstyle/.style = {block, fill=white},
    envstyle/.style = {block, fill=white}, % green!15
    advstyle/.style = {block, fill=white},
}

\begin{tikzpicture}
    [
        node distance=5.5cm,
        every node/.style={font=\large},
        align=center
    ]

    \node (system) [sutstyle] {System}; % System
    \node (environment) [envstyle, right of=system] {Environment}; % Environment
    \node (adversary) [advstyle, right of=environment] {Adversary}; % Adversary

    \draw[->] (system) -- (environment) node [pos=0.45,above] {{ action $a$}};
    \draw[->] (adversary) -- (environment) node [pos=0.45,above] {{disturbance $x$}};
    \draw[->] (environment.south) ++(-0.5,0) -- +(0,-1) -| node[pos=0.25,below] {{observation $o$}} (system);
    \draw[->] (environment.south) ++(0.5,0) -- +(0,-1) -| node[pos=0.25,below] {{state $s$}} (adversary);
\end{tikzpicture}}
\caption{Model of the safety validation problem.}
\label{fig:problem}
\end{figure}

% what is safety validation
Safety validation algorithms (\cref{fig:problem}) search for sequences of disturbances in an environment that cause an autonomous system to fail~\cite{kapinski2016simulation,corso2020survey}. At each step, the autonomous agent under test (or \emph{system}) makes an observation $o \in \mathcal{O}$ of the environment and decides to take action $a \in \mathcal{A}$. An adversary observes the state $s \in \mathcal{S}$, then applies a disturbance $x \in \mathcal{X}$ with the goal of causing the system to arrive in a set of failure states $E \subseteq \mathcal{S}$.

% It can be defined as an MDP
Safety validation can be modeled as an MDP defined by $(\mathcal{S}, \mathcal{X}, P, R, \gamma)$ where $\mathcal{S}$ represents the possible states of both the system and the environment, and $\mathcal{X}$ is the space of possible disturbances controlled by the adversary. The transition function $P(s^\prime \mid s, x)$ includes the action of the system and the dynamics of the environment. The reward function depends on the safety validation goal, and in this work we solve for the most likely failure~\cite{corso2020survey} with
\begin{equation}
    R(s, x, s^\prime) = \lambda \log p(x \mid s)  +  \mathds{1}\left\{ s^\prime \in E \right\} \label{eq:reward}
\end{equation}
where $\lambda$ is a small positive constant and $p(x \mid s)$ is the probability of $x$ occurring naturally in the environment. 

% How is safety validation a transfer learning problem
Safety validation MDPs may differ in a variety of ways, which we can explain using an example based on autonomous driving. If validation is performed on two different road geometries (e.g. highway-driving  and an intersection), then the state space and disturbance space may be different. If we fix the road geometry, then the transition model may vary if we validate different driving policies. The reward function will vary if the disturbance model changes or we alter the set of failure states. In this work, we consider systems with different behavior operating in similar environments. We, therefore, assume that the state space, disturbance space, and reward function remain fixed while the transition model varies between tasks.

\section{Proposed Approach}
In this section, we first show how to formulate iterative safety validation as a sequence of tasks and specify two ways that knowledge transfer may occur. We then introduce A2T as our choice of transfer learning algorithm and propose a modification to it.

\subsection{Problem Formulation}
Suppose we are performing safety validation on a sequence of related systems and  must validate each system before observing the next one. We model this problem as solving a sequence of MDPs (or tasks) $[T_1, T_2, \ldots]$ where the $i$th task is given by $T_i = (\mathcal{S}, \mathcal{X}, P_i, R, \gamma)$. Due to variations in system behavior, the transition model $P_i$ is unique to each task, while $\mathcal{S}$, $\mathcal{X}$, $R$, and $\gamma$ are shared across tasks. We wish to develop a learning algorithm $L$ that solves task $i$ given the $i-1$ previous solutions $[K_1, K_2, \ldots, K_{i-1}]$ such that
\begin{equation}
    K_i = L(T_i; K_{1:i-1}) \text{.}
\end{equation}
The previous solutions may take the form of value functions or policies. The learning procedure is iterative because the new solution $K_i$ can be added to the set of previous solutions when solving the next task $T_{i+1}$. Since all previous solutions are used, $L$ must avoid negative transfer by learning which source task solutions are applicable to the current task.

In the context of safety validation, we hypothesize two qualitatively distinct ways that the tasks will be related. The first case is that of a \emph{learning system} which is improving its performance from task to task. Each task is therefore more challenging for the adversary since failures that were present in previous tasks may no longer exist or may only occur due to a narrower range of disturbances. In this context, the most recent solutions are likely to provide the most relevant information and large parts of those solutions may be directly applicable to the new task. The second case is that of \emph{comparable systems} where the systems have a similar level of competency but exhibit different behavior. Disturbance trajectories that lead to failure for one system may not cause failure in any other systems, although they might share similar conceptual failure modes. In this setting, direct transfer of solutions may be ineffective, and we need to rely on other types of knowledge.

\subsection{Choice of Learning Algorithm}
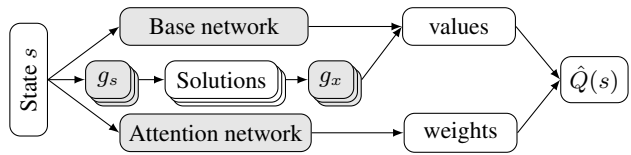
\begin{figure}
\centering
% TikZ diagram for black-box safety validation problem formulation.
\begin{tikzpicture}
    \tikzstyle{every node}=[font=\small, align=center]
    \tikzset{
        n/.style={draw, rounded corners, minimum height=0.5cm, minimum width = 1.5cm},
        n2/.style={n, minimum width=2.5cm},
        n3/.style={n, minimum width=0.6cm}
        }
    
    %state 
    \node (state) [n, rotate=90] {\small State $s$};
    
    %base network
    \node (base) [n2, fill=black!10, above right of=state, anchor=west, xshift=0.5cm] {Base network};
    
     %State Transforms
    \node (stback1) [n3, fill=black!10, right of=state, anchor=west, xshift=-0.15cm, yshift=-0.1cm] {};
    \node (stback2) [n3, fill=black!10, right of=state, anchor=west, xshift=-0.2cm, yshift=-0.05cm] {};
    \node (st) [n3, fill=black!10, right of=state, anchor=west, xshift = -0.25cm] {$g_s$};
    
    %Solutions
    \node (solutionsback1) [n, right of=st, xshift=-.15cm, yshift=-0.1cm, anchor=west] {};
    \node (solutionsback2) [n, fill=white, right of=st, xshift=-0.2cm, yshift=-0.05cm, anchor=west] {};
    \node (solutions) [n, fill=white, right of=st, anchor=west, xshift = -0.25cm] {Solutions};
    
    %Action Transforms
    \node (atback1) [n3, fill=black!10, right of=solutions, anchor=west, xshift=0.25cm, yshift=-0.1cm] {};
    \node (atback2) [n3, fill=black!10, right of=solutions, anchor=west, xshift=0.2cm, yshift=-0.05cm] {};
    \node (at) [n3, fill=black!10, right of=solutions, anchor=west, xshift=0.15cm] {$g_x$};
    
    %weights
    \node (attn) [n2, fill=black!10, below right of=state, anchor=west, xshift=0.5cm] {Attention network};

    \node (values) [n, right of=base, anchor=west, xshift = 1.5cm] {values};
    
     \node (weights) [n, right of=attn, anchor=west, xshift=1.5cm] {weights};
     
     \node (pfail) [n, minimum width = 0.5cm, right of=at, xshift=2.5cm] {$\hat{Q}(s)$};

    \draw[-latex] (state.south) -- (base.west);
    \draw[-latex] (state.south) -- (st.west);
    \draw[-latex] (st.east) + (0.1cm, 0cm) -- (solutions.west);
    \draw[-latex] (solutions.east) + (0.1cm, 0cm) -- (at.west);
    \draw[-latex] (state.south) -- (attn.west);
    
    \draw[-latex] (base.east) -- (values.west);
    \draw[-latex] (at.east) + (0.1cm, 0cm) -- (values.west);
    \draw[-latex] (attn.east) -- (weights.west);
    
    \draw[-latex] (weights.east) -- (pfail.west);
    \draw[-latex] (values.east) -- (pfail.west);

    % backprop
    
    % \draw [dashed, -latex] (weights.south) to [out=-1500,in=-60] (attn.south);
    % \draw [dashed, -latex] (pfail.south) to [out=-150,in=-60] (weights.south);
    % \draw [dashed, -latex] (values.north) to [out=150,in=60] (base.north);
    % \draw [dashed, -latex] (pfail.north) to [out=150,in=60] (values.north);
\end{tikzpicture}
\caption{A2T network with state and action transformation.}
\label{fig:a2t_architecture}
\end{figure}

To accelerate safety validation we use the attend, adapt, and transfer (A2T) learning algorithm~\cite{rajendran2017attend} with a minor modification. A2T accelerates learning on a new task by combining the solutions of $k$ previous tasks using a learned set of state-dependent attention weights. To avoid negative transfer, A2T simultaneously learns a solution from scratch so there are a total of $k+1$ solutions and as  many attention weights. A2T can be used to estimate optimal policies or optimal value functions and we found it most straightforward to estimate the optimal action value function. When $Q^*$ is estimated by a neural network, called a $Q$-network, the input is a vector representing the state and the output is a vector representing the values of a discrete set of disturbances. To make this clear, if $\mathcal{X} \subseteq \mathbb{R}^m$, then $\hat{Q}(s) \in \mathbb{R}^m$, is a vector that represents the estimates of the optimal values for each disturbance. The action value function is estimated by the expression
\begin{equation}
\hat{Q}(s) = w_0(s) \hat{Q}_{\rm base}(s) + \sum_{i=1}^k w_i(s) \hat{Q}_i(s) \label{eq:A2T}
\end{equation}
where $\hat{Q}_i$ comes from the $i$th source task, $\hat{Q}_{\rm base}$ is learned from scratch and $w_i(s)$ is the $i$th attention weight, normalized so $\sum_{i=0}^m w_i(s) = 1$.

To handle substantially different system behaviors (as in the comparable systems setting), we propose a modification of A2T where we include a learned transformation of the state and action value function for each of the $k$ previous solutions. If the state space $\mathcal{S} \subseteq \mathbb{R}^n$, then we define a \emph{state transformation} as a function $g_s : \mathbb{R}^n \to \mathbb{R}^n$ and an \emph{action value transformation} as a function $g_x : \mathbb{R}^m \to \mathbb{R}^m$. Applied to the A2T algorithm, the action value estimate with state and action space transformations is given by
\begin{equation}
\hat{Q}(s) = w_0(s) \hat{Q}_{\rm base}(s) + \sum_{i=1}^k w_i(s) g_x^{(i)}(\hat{Q}_i(g_s^{(i)}(s)))
\end{equation}
where $g_s^{(i)}$ and $g_x^{(i)}$ are the state and action value transformations for the $i$th source solution. In this work, we use linear transformations because they are effective and simple.

The A2T algorithm with state and action value transformations can be encoded as the network architecture shown in \cref{fig:a2t_architecture}. The state is used as input to the base network, the source solutions, and the attention network. The base network is a $Q$-network that learns from scratch. The $k$ source solutions are the $Q$-networks that represent the optimal solutions of the source tasks. The source solutions are preceded by a state transformation and followed by an action value transformation. These transformations have the same output dimension as input dimension and are initialized to the identity transformation (with a small amount of noise for breaking symmetry). The attention network has $k+1$ output units with a softmax layer for normalization. The $Q$ values of each source solution and the base network are weighted by the corresponding attention weights and summed together to get a final estimate. The network parameters from the base network, attention network, and state and action value transformations are trained using DQN. If the source solutions are not differentiable with respect to the state then we would apply gradient free optimization~\cite{kochenderfer2019algorithms} to the state transformation.

\begin{figure}
    \centering
    \captionsetup[subfigure]{format=hang}
    \begin{subfigure}[t]{0.25\linewidth}
        \centering
        \includegraphics[width=\textwidth]{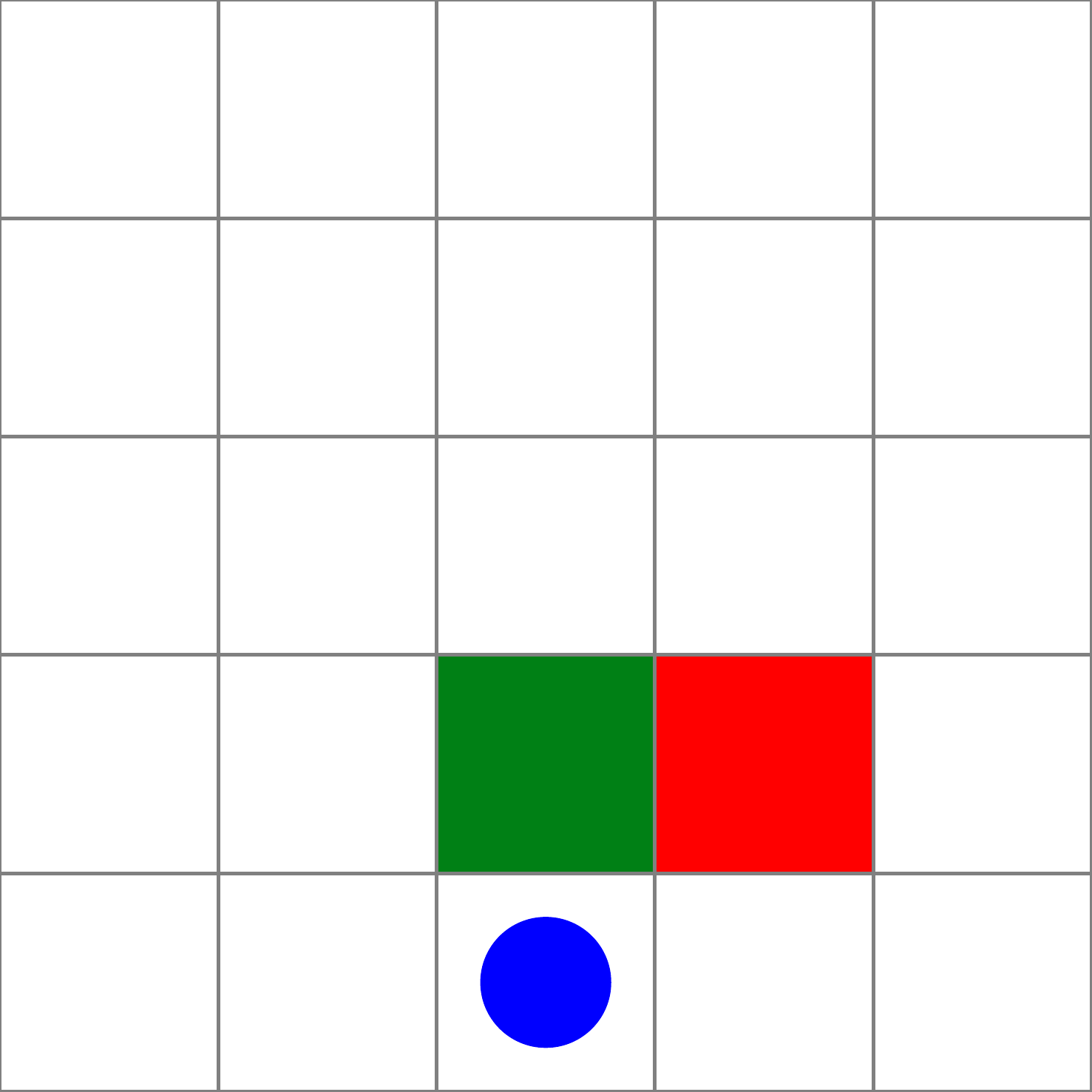}
        \caption{\small Source MDP}    
        \label{fig:motiation_source}
    \end{subfigure}
    \hfill
    \begin{subfigure}[t]{0.25\linewidth}  
        \centering 
        \includegraphics[width=\textwidth]{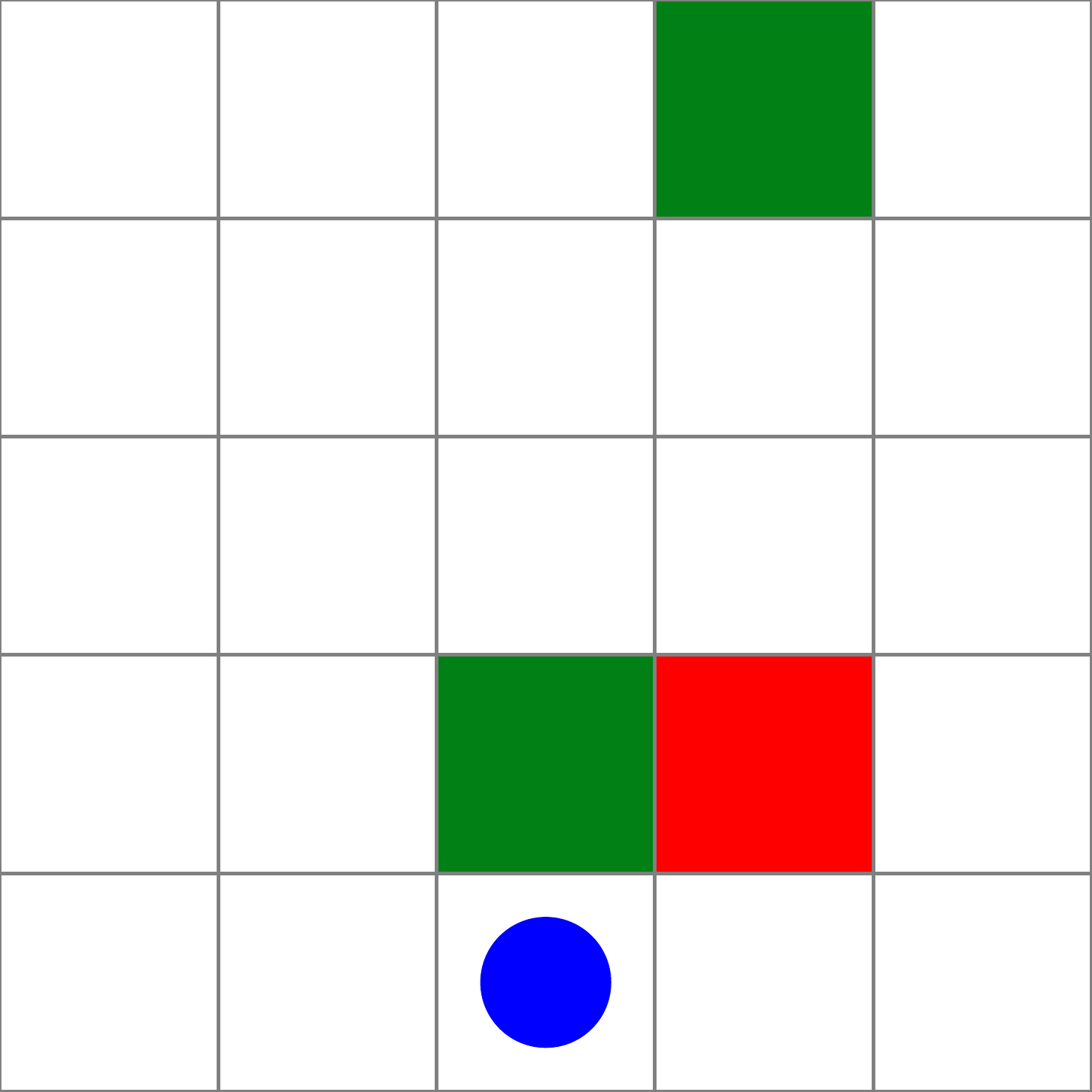}
        \caption{\small Similar}  
        \label{fig:motivation_a2t}
    \end{subfigure}
    \hfill
    \begin{subfigure}[t]{0.25\linewidth}   
        \centering 
        \includegraphics[width=\textwidth]{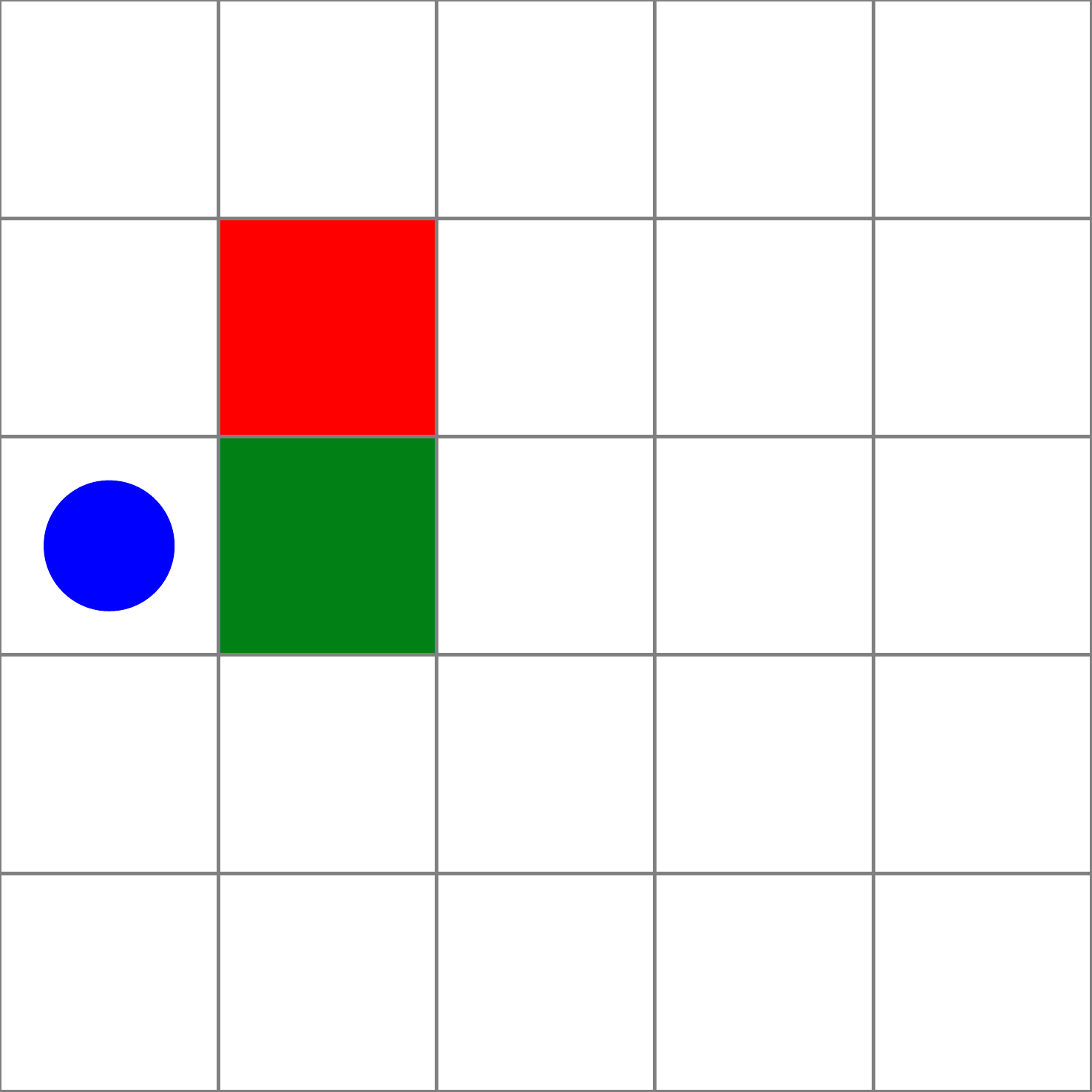}
        \caption{\small Transformed}
        \label{fig:motivation_sat}
    \end{subfigure}
    \caption{Knowledge transfer scenarios.}
    \label{fig:motivation}
\end{figure}

\paragraph{Motivation.} Here we provide some intuition for our choice of transfer learning algorithm and the reason for the state and action transformations. Suppose we are solving a sequence of tasks (\cref{fig:motivation}), each is a gridworld where an agent (blue circle) moves between adjacent squares attempting to achieve high reward (green squares) while avoiding states with low reward (red squares). Given the optimal action value function for a source MDP (\cref{fig:motiation_source}), we wish to transfer it to two related tasks. In the first transfer problem (\cref{fig:motivation_a2t}), the reward distribution is locally similar to the source MDP, and only differs in one state. In this case, a policy that works well in the source task will also work well in the new task, especially when the agent is in a state where the local reward landscape matches up (as depicted). A2T will work well in this setting because it can quickly learn attention weights that favor the source policy in most states, and only require our baseline solution when the agent is near the top right corner. In the second transfer problem (\cref{fig:motivation_sat}), A2T is likely to behave poorly because there are no states in which the source policy can be directly applied to achieve high reward. If, however, we could transform the state space by reflecting it across the diagonal and rotate the actions of the agent by \SI{90}{\degree}, then we could directly apply the source policy. This is the motivation for applying transformations both before and after the source solutions.

\section{Experiments}

This section describes two safety validation problems: a gridworld scenario (GW) and an autonomous driving scenario (AD). Each scenario has two transfer learning problems, one for validating a learning system that improves over time, and the other for validating a set of comparable systems with different behaviors. We then describe the experimental setup and how we compute the evaluation metrics. 

% Gridworld with adversary

\subsection{Gridworld with Adversary}
\begin{figure}
\centering
\includegraphics[width=0.45\linewidth]{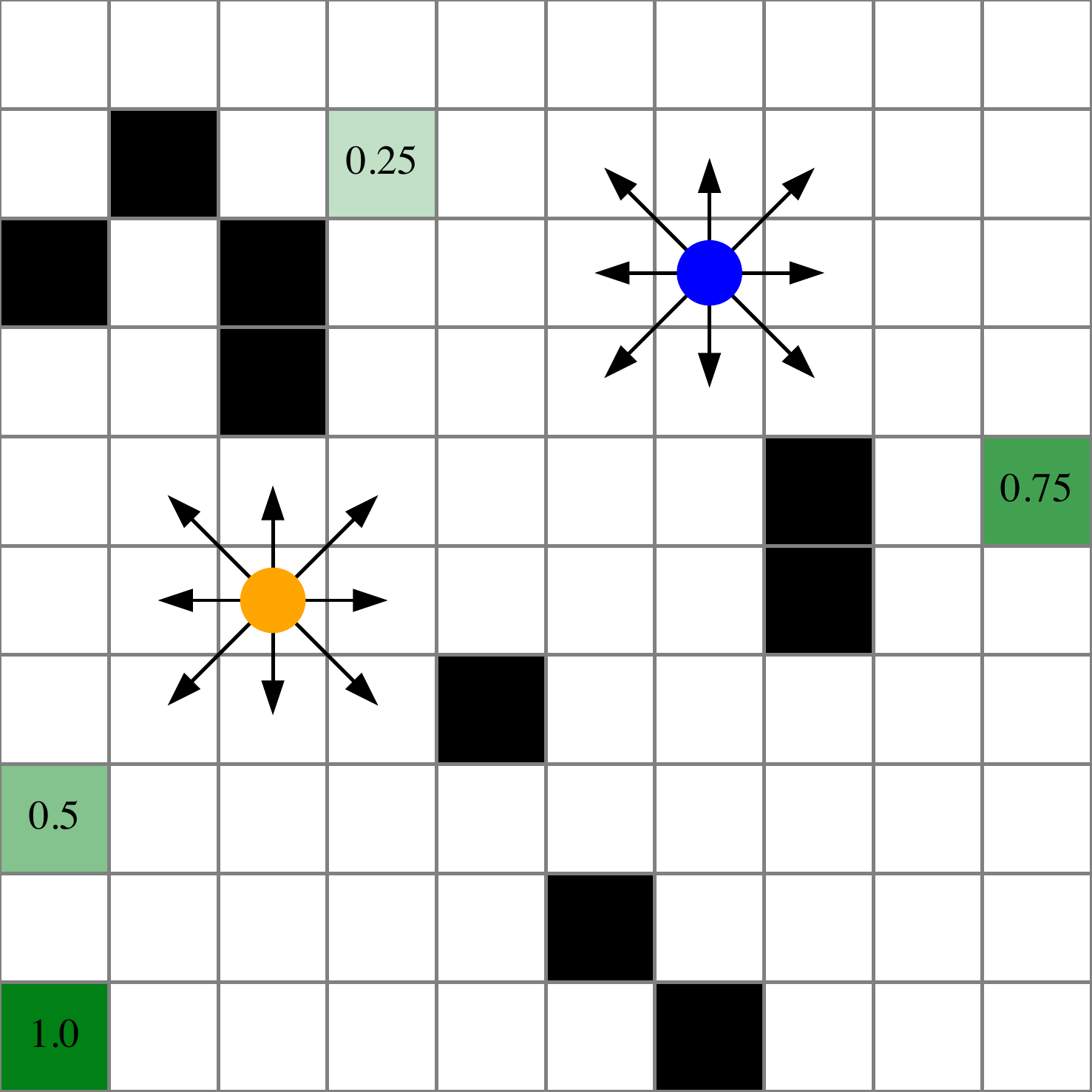}
\caption{Gridworld with adversary scenario. }
\label{fig:gridworld_with_adv}
\end{figure}

% Description of the MDP
The first safety validation problem we model is a gridworld with two agents, shown in \cref{fig:gridworld_with_adv}. The system is the agent in blue who is trying to arrive at one of the squares with positive reward while avoiding the adversary (orange agent). Both agents are initialized randomly and can move to any adjacent or diagonal state that is not a wall (marked as black). As a safety validation MDP, the state is the grid location of both agents and the disturbances are the actions of the adversary: (up, down, left, right, up right, up left, down right, down left, stay). At each step, the blue agent chooses an action based on its policy and transitions to the appropriate state with probability \num{0.7} and transitions to a random feasible state with probability \num{0.3}.  Meanwhile, the orange agent transitions to the state specified by the disturbance $x$. The episode ends when the blue agent arrives at a square with positive reward or when the two agents collide in the same state, which is a failure of the system. We model each disturbance as equally likely to happen so we give a reward of \num{1} for finding a failure and \num{0} otherwise.  

% Description of the difference between tasks
We design two sets of tasks that correspond to the learning system and comparable systems settings. For the learning system, the blue agent is trained using DQN against an orange agent that behaves randomly. Over $\num{e6}$ training steps, \num{10} versions of the system policy were stored, each with an increasing level of performance. Each safety validation task has the adversary validate an increasingly capable version of the learning system. For the comparable systems setting, each task has a different distribution of reward locations, reward values and location of walls. The system learns an optimal policy using dynamic programming~\cite{dmubook}, assuming the adversary behaves randomly. Each system is therefore equally competent, but some configurations of the gridworld are more challenging than others.

% Autonomous driving with blind spot
\subsection{Autonomous Vehicle}
\begin{figure}
\centering
\includegraphics[trim={0 7cm 10cm 6.5cm},clip, width=0.7\linewidth]{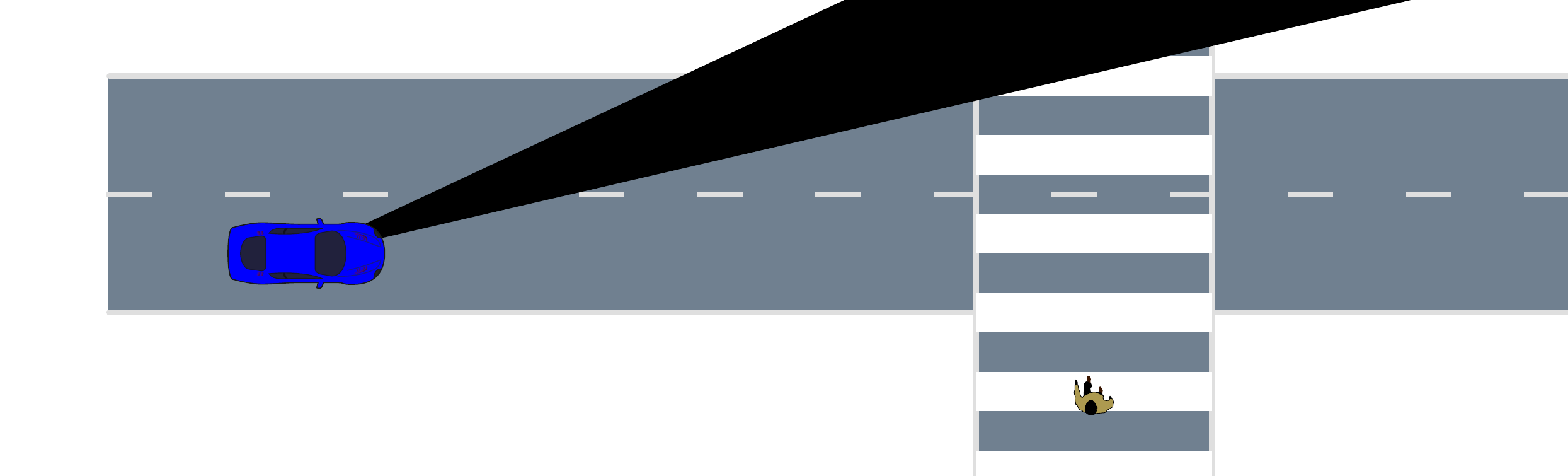}
\caption{Autonomous driving scenario}
\label{fig:av_blindspot}
\end{figure}

% Description of the MDP
The second safety validation problem we model is an autonomous vehicle navigating an intersection with a crossing pedestrian, shown in \cref{fig:av_blindspot}. The system is a vehicle controlled by the intelligent driver model (IDM)~\cite{kesting2010enhanced}, a rule-based driving policy that avoids collisions, and we add a blind spot with a specified direction and angular width. The vehicle tries to reach the end of the road while yielding to the pedestrian. Both agents are initialized randomly in a starting range of initial conditions. 

As a safety validation MDP, the state is the position and velocity of the agents and the disturbances are accelerations of the pedestrian. The pedestrian can accelerate up, down, left, or right by \SI{1}{m/s^2} with probability \num{0.01} or have no acceleration with probability \num{0.96}. The velocity of the pedestrian is limited to an absolute value of \SI{3}{m/s}. At each step, the vehicle chooses an acceleration based on the IDM rules and the location of the pedestrian.  The pedestrian acceleration is specified by the choice of disturbance. Both agents have their position and velocity updated deterministically from their current state and acceleration. The episode terminates when the vehicle reaches the end of the road safely or a collision occurs between the pedestrian and the vehicle. The reward function is given by \cref{eq:reward} where the set $E$ defined by any state where the vehicle and pedestrian overlap.  

% Description of the difference between tasks
For the autonomous driving scenario, we also design two sets of tasks corresponding to a learning system setting and a comparable systems setting. Since the autonomous vehicle does not use machine learning, we simulate an improvement by progressively shrinking the blind spot of the vehicle. The blind spot remains in the same direction (\SI{20}{\degree} from the horizontal), but reduces in width from \SI{30}{\degree} to \SI{6}{\degree} over \num{10} iterations. The vehicle therefore has a decreasing rate of failures over the tasks but there is some overlap in failure modes between adjacent tasks. For the comparable systems setting, each system has a blind spot sampled uniformly at random with a direction in the range $[\SI{-30}{\degree}, \SI{30}{\degree}]$ and an angular width in the range $[\SI{3}{\degree}, \SI{9}{\degree}]$. From a population of \num{30} tasks, we selected \num{9} tasks that differed substantially, to make the transfer problem as challenging as possible within our setting. Two tasks differed if the optimal safety validation policy of one performs poorly on the other.

\subsection{Experimental Setup}

The experimental procedure is as follows. For each task in a set of tasks, we solve for an optimal $Q$-network from scratch using DQN with prioritized replay, double $Q$-learning~\cite{Hasselt2016deep}, and the Huber loss. We construct a learning curve during training by periodically storing the evaluation of the $Q$-network. For the second task onward, we then solve it using the same learning algorithm with the following $Q$-network  architectures:
\begin{itemize}
    \item \textbf{Fine-tune}: Train the last layer of the previous $Q$-network.
    \item \textbf{A2T}: A2T architecture with previous $Q$-networks as the source solutions.
    \item \textbf{A2T+SAVT}:  A2T architecture augmented with linear state and action value transformations.
\end{itemize}
The networks are initialized with Xavier initialization~\cite{glorot2010understanding}, while the transformations were initialized to the identity matrix with uniform random noise  in the range [-\num{1e-3}, \num{1e-3}] added to the parameters to break symmetry. Additional information on network architecture and hyperparameters is shown in \cref{tab:hyperparameters}.

\begin{table}
\small
    \centering
    \caption{Network architectures and hyperparameters. }
    \label{tab:hyperparameters}
    \begin{tabular}{@{}ll@{}} 
        \toprule
        \textbf{Parameter} & \textbf{Value}  \\
        \midrule
        Base network & 3 hidden layers, [\num{64}, \num{32}, \num{16}] relus\\
        Attention network & 1 hidden layer, \num{16} relus \\
        Training steps & \num{3e6} \\
        Batch size & \num{64} \\
        Learning rate $\alpha$ & \num{4e-5} (GW), \num{5e-5} (AD) \\
        Target update frequency & \num{2000} (GW), \num{3000} (AD) \\
        Evaluation & \num{300} episodes every \num{2000} steps \\
        Exploration policy & $\epsilon$-greedy with $\epsilon \in [1, 0.1]$\\
        \bottomrule
    \end{tabular}
\end{table}

% Processing the learning curves
We filter each learning curve using a moving-average filter with a width of \num{20} evaluations steps to help remove the noise due to finite sample evaluation and any outlier evaluation points. For each learning curve we identify the \emph{near-optimal} performance as $\mu - \sigma$, where $\mu$ and $\sigma$ are the mean and standard deviation of the performance in a window with a width of \num{100} evaluation steps around the point of maximum performance. We use near-optimal performance because it is a more stable measure of how fast the learning took place than the point of maximum performance.

% computing evaluation metrics
The jumpstart is the difference in initial performance between a transfer and no-transfer learning algorithms. It can be computed from the first entries in the learning curves. When reporting jumpstart, we only include fine-tuning and A2T because A2T+SAVT has the same outputs as A2T until the transformations deviate from identity. The final performance is the difference in near-optimal performance between the transfer and no-transfer learning algorithms. The steps to threshold metric measures how many training steps are required for a transfer learning algorithm to reach the near-optimal performance of the no-transfer algorithm.

% normalization of metrics. 
The metrics are normalized with reference to the learning curve of the no-transfer algorithm because the initial and near-optimal performance varies between tasks. Let $y$ be the performance (initial or final) of a transfer learning algorithm and $y_{\rm ref}$ be the performance of the no-transfer learning algorithm, then we report the fractional difference in performance $(y - y_{\rm ref})/|y_{\rm ref}|$. Let $t$ be the number of training steps required to reach a threshold for a transfer learning algorithm and $t_{\rm ref}$ be the same quantity for the no-transfer learning algorithm, then we report the ratio $t / t_{\rm ref}$.

% Notes
We use the number of training steps rather than the wall clock time because we assume that the cost of running the simulator is much larger than the cost of updating the parameters of the model. This is a good assumption for high-fidelity simulators that are often used for validating safety-critical systems. We also assume that the training time of previous source tasks is a sunk cost and it is not included in our efficiency metric. This assumption is valid in the case of iterative safety validation because the new version of the system must be validated regardless of the approach used. When using A2T in a real-world setting, we would not solve each task from scratch and therefore the source solutions would take the form of A2T networks. A practitioner may wish to compress the A2T network~\cite{julian2019deep} into a traditional architecture before it is used as a source solution. We chose to use the networks trained from scratch for ease of implementation and to isolate the effects of transfer learning from other issues.

\section{Results and Discussion}
\begin{figure*}
    \centering
    \begin{subfigure}[b]{0.32\textwidth}
        \centering
        \begin{tikzpicture}[/tikz/background rectangle/.style={fill={rgb,1:red,1.0;green,1.0;blue,1.0}, draw opacity={1.0}}, show background rectangle]
\begin{axis}[point meta max={nan}, point meta min={nan},  axis background/.style={fill={rgb,1:red,1.0;green,1.0;blue,1.0}, opacity={0.0}}, anchor={north west}, xshift={1.0mm}, yshift={-1.0mm}, width=\textwidth, height={35mm}, scaled x ticks={false}, xlabel={Task Iteration}, x tick style={color={rgb,1:red,0.0;green,0.0;blue,0.0}, opacity={1.0}}, x tick label style={color={rgb,1:red,0.0;green,0.0;blue,0.0}, opacity={1.0}, rotate={0}}, xlabel style={at={(ticklabel cs:0.5)}, anchor=near ticklabel, font={{\fontsize{8 pt}{10.4 pt}\selectfont}}, color={rgb,1:red,0.0;green,0.0;blue,0.0}, draw opacity={1.0}, rotate={0.0}}, xmajorgrids={true}, xmin={1.76}, xmax={10.24}, xtick={{2.0,4.0,6.0,8.0,10.0}}, xticklabels={{$2$,$4$,$6$,$8$,$10$}}, xtick align={inside}, xticklabel style={font={{\fontsize{8 pt}{10.4 pt}\selectfont}}, color={rgb,1:red,0.0;green,0.0;blue,0.0}, draw opacity={1.0}, rotate={0.0}}, x grid style={color={rgb,1:red,0.0;green,0.0;blue,0.0}, draw opacity={0.1}, line width={0.5}, solid}, axis x line*={left}, x axis line style={color={rgb,1:red,0.0;green,0.0;blue,0.0}, draw opacity={1.0}, line width={1}, solid}, scaled y ticks={false}, ylabel={Jumpstart}, y tick style={color={rgb,1:red,0.0;green,0.0;blue,0.0}, opacity={1.0}}, y tick label style={color={rgb,1:red,0.0;green,0.0;blue,0.0}, opacity={1.0}, rotate={0}}, ylabel style={at={(ticklabel cs:0.5)}, anchor=near ticklabel, font={{\fontsize{8 pt}{10.4 pt}\selectfont}}, color={rgb,1:red,0.0;green,0.0;blue,0.0}, draw opacity={1.0}, rotate={0.0}}, ymajorgrids={true}, ymin={0}, ymax={5.399999826954267}, ytick={{0.0,1.0,2.0,3.0,4.0,5.0}}, yticklabels={{$0$,$1$,$2$,$3$,$4$,$5$}}, ytick align={inside}, yticklabel style={font={{\fontsize{8 pt}{10.4 pt}\selectfont}}, color={rgb,1:red,0.0;green,0.0;blue,0.0}, draw opacity={1.0}, rotate={0.0}}, y grid style={color={rgb,1:red,0.0;green,0.0;blue,0.0}, draw opacity={0.1}, line width={0.5}, solid}, axis y line*={left}, y axis line style={color={rgb,1:red,0.0;green,0.0;blue,0.0}, draw opacity={1.0}, line width={1}, solid}]
    \addplot[color={rgb,1:red,0.192;green,0.639;blue,0.329}, name path={7109be5d-bad6-461b-9b58-22c0764bd826}, draw opacity={1.0}, line width={1}, solid, mark={square*}, mark size={1.125 pt}, mark repeat={1}, mark options={color={rgb,1:red,0.0;green,0.0;blue,0.0}, draw opacity={1.0}, fill={rgb,1:red,0.192;green,0.639;blue,0.329}, fill opacity={1.0}, line width={0.75}, rotate={0}, solid}]
        table[row sep={\\}]
        {
            \\
            2.0  5.399999826954267  \\
            3.0  1.64337356431787  \\
            4.0  4.311526759144857  \\
            5.0  2.5137502909917884  \\
            6.0  2.419672023023025  \\
            7.0  2.641755706853315  \\
            8.0  2.296500713856086  \\
            9.0  2.479999886751185  \\
            10.0  1.7107280148477237  \\
        }
        ;
    \addplot[color={rgb,1:red,0.87;green,0.176;blue,0.149}, name path={4a4f8dfc-4212-4019-a490-23ef38683db4}, draw opacity={1.0}, line width={1}, dotted, mark={triangle*}, mark size={1.125 pt}, mark repeat={1}, mark options={color={rgb,1:red,0.0;green,0.0;blue,0.0}, draw opacity={1.0}, fill={rgb,1:red,0.87;green,0.176;blue,0.149}, fill opacity={1.0}, line width={0.75}, rotate={0}, solid}]
        table[row sep={\\}]
        {
            \\
            2.0  5.0741933581298255  \\
            3.0  0.8746987900543249  \\
            4.0  3.087227503701616  \\
            5.0  3.030000289455106  \\
            6.0  0.9327868882493604  \\
            7.0  1.1174377266660267  \\
            8.0  0.8987108727254944  \\
            9.0  2.8999998062849217  \\
            10.0  1.192848061467238  \\
        }
        ;
\end{axis}
\end{tikzpicture}
        \caption{Jumpstart improvement.}
        \label{fig:gwl_jumpstart}
    \end{subfigure}
    \hfill
    \begin{subfigure}[b]{0.32\textwidth}
        \centering
        \begin{tikzpicture}[/tikz/background rectangle/.style={fill={rgb,1:red,1.0;green,1.0;blue,1.0}, draw opacity={0.0}}, show background rectangle]
\begin{axis}[point meta max={nan}, point meta min={nan}, legend cell align={left}, title={}, title style={at={{(0.5,1)}}, anchor={south}, font={{\fontsize{14 pt}{18.2 pt}\selectfont}}, color={rgb,1:red,0.0;green,0.0;blue,0.0}, draw opacity={1.0}, rotate={0.0}}, legend style={color={rgb,1:red,0.0;green,0.0;blue,0.0}, draw opacity={0.0}, line width={1}, solid, fill={rgb,1:red,1.0;green,1.0;blue,1.0}, fill opacity={1.0}, text opacity={1.0}, font={{\fontsize{8 pt}{10.4 pt}\selectfont}},at={(0.01, 1.15)}, anchor={south west}, legend columns=-1}, legend style={/tikz/every even column/.append style={column sep=0.5cm}}, axis background/.style={fill={rgb,1:red,1.0;green,1.0;blue,1.0}, opacity={1.0}}, anchor={north west}, xshift={1.0mm}, yshift={-1.0mm}, width=\textwidth, height={35mm}, scaled x ticks={false}, xlabel={Task Iteration}, x tick style={color={rgb,1:red,0.0;green,0.0;blue,0.0}, opacity={1.0}}, x tick label style={color={rgb,1:red,0.0;green,0.0;blue,0.0}, opacity={1.0}, rotate={0}}, xlabel style={at={(ticklabel cs:0.5)}, anchor=near ticklabel, font={{\fontsize{8 pt}{10.4 pt}\selectfont}}, color={rgb,1:red,0.0;green,0.0;blue,0.0}, draw opacity={1.0}, rotate={0.0}}, xmajorgrids={true}, xmin={1.76}, xmax={10.24}, xtick={{2.0,4.0,6.0,8.0,10.0}}, xticklabels={{$2$,$4$,$6$,$8$,$10$}}, xtick align={inside}, xticklabel style={font={{\fontsize{8 pt}{10.4 pt}\selectfont}}, color={rgb,1:red,0.0;green,0.0;blue,0.0}, draw opacity={1.0}, rotate={0.0}}, x grid style={color={rgb,1:red,0.0;green,0.0;blue,0.0}, draw opacity={0.1}, line width={0.5}, solid}, axis x line*={left}, x axis line style={color={rgb,1:red,0.0;green,0.0;blue,0.0}, draw opacity={1.0}, line width={1}, solid}, scaled y ticks={false}, ylabel={Final Improvement}, y tick style={color={rgb,1:red,0.0;green,0.0;blue,0.0}, opacity={1.0}}, y tick label style={color={rgb,1:red,0.0;green,0.0;blue,0.0}, opacity={1.0}, rotate={0}}, ylabel style={at={(ticklabel cs:0.5)}, anchor=near ticklabel, font={{\fontsize{8 pt}{10.4 pt}\selectfont}}, color={rgb,1:red,0.0;green,0.0;blue,0.0}, draw opacity={1.0}, rotate={0.0}}, ymajorgrids={true}, ymin={-0.9999999992516994}, ymax={1.4}, ytick={{-0.5,0.0,0.5,1.0}}, yticklabels={{$-0.5$,$0.0$,$0.5$,$1.0$}}, ytick align={inside}, yticklabel style={font={{\fontsize{8 pt}{10.4 pt}\selectfont}}, color={rgb,1:red,0.0;green,0.0;blue,0.0}, draw opacity={1.0}, rotate={0.0}}, y grid style={color={rgb,1:red,0.0;green,0.0;blue,0.0}, draw opacity={0.1}, line width={0.5}, solid}, axis y line*={left}, y axis line style={color={rgb,1:red,0.0;green,0.0;blue,0.0}, draw opacity={1.0}, line width={1}, solid}]
    \addplot[color={rgb,1:red,0.192;green,0.51;blue,0.74}, name path={48fdfafb-9d23-413b-a6ad-a374fc34b6af}, draw opacity={1.0}, line width={1}, dashdotted, mark={*}, mark size={1.125 pt}, mark repeat={1}, mark options={color={rgb,1:red,0.0;green,0.0;blue,0.0}, draw opacity={1.0}, fill={rgb,1:red,0.192;green,0.51;blue,0.74}, fill opacity={1.0}, line width={0.75}, rotate={0}, solid}]
        table[row sep={\\}]
        {
            \\
            2.0  0.03367594816657809  \\
            3.0  -0.07097311961060898  \\
            4.0  0.06687817125438633  \\
            5.0  0.16678621875512595  \\
            6.0  0.09843160169116001  \\
            7.0  0.5865169076602881  \\
            8.0  -0.08001339757544591  \\
            9.0  0.28600602437414385  \\
            10.0  0.17532166490692389  \\
        }
        ;
    \addlegendentry {A2T w/ SAVT}
    \addplot[color={rgb,1:red,0.192;green,0.639;blue,0.329}, name path={0708b760-43cd-4e22-851e-b0e88496cb80}, draw opacity={1.0}, line width={1}, solid, mark=square*, mark size={1.125 pt}, mark repeat={1}, mark options={color={rgb,1:red,0.0;green,0.0;blue,0.0}, draw opacity={1.0}, fill={rgb,1:red,0.192;green,0.639;blue,0.329}, fill opacity={1.0}, line width={0.75}, rotate={0}, solid}]
        table[row sep={\\}]
        {
            \\
            2.0  0.0979433272941114  \\
            3.0  0.06459659586256183  \\
            4.0  0.028538583298642704  \\
            5.0  0.3141625774710388  \\
            6.0  0.10870726139269066  \\
            7.0  0.6544941263643854  \\
            8.0  0.004687056164156816  \\
            9.0  0.23378228208335256  \\
            10.0  -0.09575985193285183  \\
        }
        ;
    \addlegendentry {A2T}
    \addplot[color={rgb,1:red,0.87;green,0.176;blue,0.149}, name path={7cb74daf-ee2a-4f82-a9c7-ba15682efcd2}, draw opacity={1.0}, line width={1}, dotted, mark=triangle*, mark size={1.125 pt}, mark repeat={1}, mark options={color={rgb,1:red,0.0;green,0.0;blue,0.0}, draw opacity={1.0}, fill={rgb,1:red,0.87;green,0.176;blue,0.149}, fill opacity={1.0}, line width={0.75}, rotate={0}, solid}]
        table[row sep={\\}]
        {
            \\
            2.0  -0.21208241547548554  \\
            3.0  -0.4826726471591091  \\
            4.0  -0.6137214993031977  \\
            5.0  -0.4162473345026299  \\
            6.0  -0.7030827656063745  \\
            7.0  -0.6224719176748725  \\
            8.0  -0.9999999992516994  \\
            9.0  -0.06976735363519794  \\
            10.0  -0.827060504880624  \\
        }
        ;
    \addlegendentry {Fine-tune}
\end{axis}
\end{tikzpicture}
        \caption{Final improvement.}
        \label{fig:gwl_final}
    \end{subfigure}
    \hfill
    \begin{subfigure}[b]{0.32\textwidth}
        \centering
        \begin{tikzpicture}[/tikz/background rectangle/.style={fill={rgb,1:red,1.0;green,1.0;blue,1.0}, draw opacity={0.0}}, show background rectangle]
\begin{axis}[point meta max={nan}, point meta min={nan}, axis background/.style={fill={rgb,1:red,1.0;green,1.0;blue,1.0}, opacity={1.0}}, anchor={north west}, xshift={1.0mm}, yshift={-1.0mm}, width=\textwidth, height={35mm}, scaled x ticks={false}, xlabel={Task Iteration}, x tick style={color={rgb,1:red,0.0;green,0.0;blue,0.0}, opacity={1.0}}, x tick label style={color={rgb,1:red,0.0;green,0.0;blue,0.0}, opacity={1.0}, rotate={0}}, xlabel style={at={(ticklabel cs:0.5)}, anchor=near ticklabel, font={{\fontsize{8 pt}{10.4 pt}\selectfont}}, color={rgb,1:red,0.0;green,0.0;blue,0.0}, draw opacity={1.0}, rotate={0.0}}, xmajorgrids={true}, xmin={1.76}, xmax={10.24}, xtick={{2.0,4.0,6.0,8.0,10.0}}, xticklabels={{$2$,$4$,$6$,$8$,$10$}}, xtick align={inside}, xticklabel style={font={{\fontsize{8 pt}{10.4 pt}\selectfont}}, color={rgb,1:red,0.0;green,0.0;blue,0.0}, draw opacity={1.0}, rotate={0.0}}, x grid style={color={rgb,1:red,0.0;green,0.0;blue,0.0}, draw opacity={0.1}, line width={0.5}, solid}, axis x line*={left}, x axis line style={color={rgb,1:red,0.0;green,0.0;blue,0.0}, draw opacity={1.0}, line width={1}, solid}, scaled y ticks={false}, ylabel={Step Ratio}, y tick style={color={rgb,1:red,0.0;green,0.0;blue,0.0}, opacity={1.0}}, y tick label style={color={rgb,1:red,0.0;green,0.0;blue,0.0}, opacity={1.0}, rotate={0}}, ylabel style={at={(ticklabel cs:0.5)}, anchor=near ticklabel, font={{\fontsize{8 pt}{10.4 pt}\selectfont}}, color={rgb,1:red,0.0;green,0.0;blue,0.0}, draw opacity={1.0}, rotate={0.0}}, ymajorgrids={true}, ymin={-0.1}, ymax={1.55408795660116}, ytick={{0.0,0.5,1.0,1.5}}, yticklabels={{$0.0$,$0.5$,$1.0$,$1.5$}}, ytick align={inside}, yticklabel style={font={{\fontsize{8 pt}{10.4 pt}\selectfont}}, color={rgb,1:red,0.0;green,0.0;blue,0.0}, draw opacity={1.0}, rotate={0.0}}, y grid style={color={rgb,1:red,0.0;green,0.0;blue,0.0}, draw opacity={0.1}, line width={0.5}, solid}, axis y line*={left}, y axis line style={color={rgb,1:red,0.0;green,0.0;blue,0.0}, draw opacity={1.0}, line width={1}, solid}]
    \addplot[color={rgb,1:red,0.192;green,0.51;blue,0.74}, name path={cff93711-456e-437e-bfc7-abc49790df81}, draw opacity={1.0}, line width={1}, dashdotted, mark={*}, mark size={1.125 pt}, mark repeat={1}, mark options={color={rgb,1:red,0.0;green,0.0;blue,0.0}, draw opacity={1.0}, fill={rgb,1:red,0.192;green,0.51;blue,0.74}, fill opacity={1.0}, line width={0.75}, rotate={0}, solid}]
        table[row sep={\\}]
        {
            \\
            2.0  0.5816671850254768  \\
            3.0  1.0835485732951422  \\
            4.0  0.7893455320488469  \\
            5.0  0.6568164147327786  \\
            6.0  1.0602784077124818  \\
            7.0  0.7162831656719646  \\
        }
        ;
    \addplot[color={rgb,1:red,0.192;green,0.51;blue,0.74}, name path={cff93711-456e-437e-bfc7-abc49790df81}, draw opacity={1.0}, line width={1}, dashdotted, mark={*}, mark size={1.125 pt}, mark repeat={1}, mark options={color={rgb,1:red,0.0;green,0.0;blue,0.0}, draw opacity={1.0}, fill={rgb,1:red,0.192;green,0.51;blue,0.74}, fill opacity={1.0}, line width={0.75}, rotate={0}, solid}, forget plot]
        table[row sep={\\}]
        {
            \\
            9.0  0.511680261715557  \\
            10.0  1.0364403459103877  \\
        }
        ;
    \addplot[color={rgb,1:red,0.192;green,0.639;blue,0.329}, name path={00265a56-bab5-4f3b-aae2-f3e046345940}, draw opacity={1.0}, line width={1}, solid, mark={square*}, mark size={1.125 pt}, mark repeat={1}, mark options={color={rgb,1:red,0.0;green,0.0;blue,0.0}, draw opacity={1.0}, fill={rgb,1:red,0.192;green,0.639;blue,0.329}, fill opacity={1.0}, line width={0.75}, rotate={0}, solid}]
        table[row sep={\\}]
        {
            \\
            2.0  0.5235527793782135  \\
            3.0  0.8146236175301722  \\
            4.0  0.7787331939262042  \\
            5.0  0.3200874741964534  \\
            6.0  0.34501570896815875  \\
            7.0  0.6744436632219665  \\
            8.0  0.5811664016916935  \\
            9.0  0.4415312482296162  \\
            10.0  1.55408795660116  \\
        }
        ;
\end{axis}
\end{tikzpicture}
        \caption{Step ratio to threshold.}
        \label{fig:gwl_step}
    \end{subfigure}
    \caption{Evaluation metrics for the gridworld scenario with a learning system.}
    \label{fig:gwl}
    \vskip -0.5 in
\end{figure*}
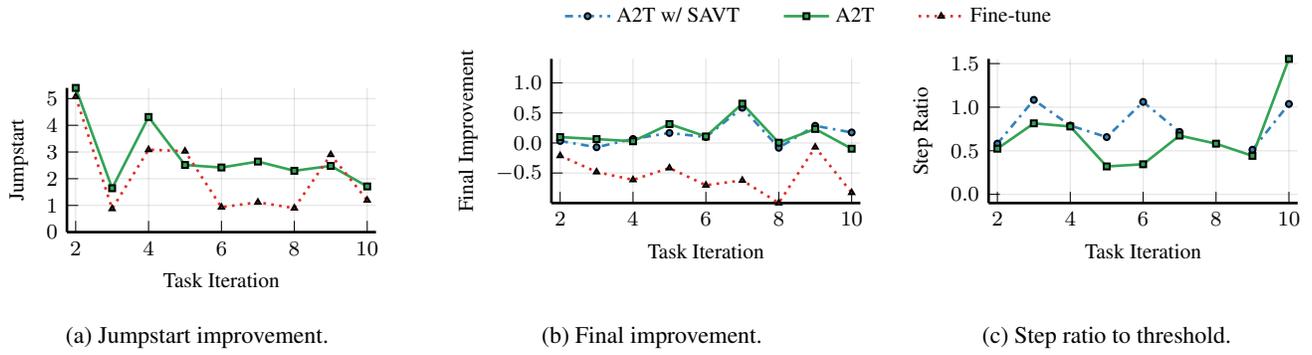

\begin{figure*}
    \centering
    \begin{subfigure}[b]{0.32\textwidth}
        \centering
        \begin{tikzpicture}[/tikz/background rectangle/.style={fill={rgb,1:red,1.0;green,1.0;blue,1.0}, draw opacity={.0}}, show background rectangle]
\begin{axis}[point meta max={nan}, point meta min={nan}, axis background/.style={fill={rgb,1:red,1.0;green,1.0;blue,1.0}, opacity={1.0}}, anchor={north west}, xshift={1.0mm}, yshift={-1.0mm}, width=\textwidth, height={35mm}, scaled x ticks={false}, xlabel={Task Iteration}, x tick style={color={rgb,1:red,0.0;green,0.0;blue,0.0}, opacity={1.0}}, x tick label style={color={rgb,1:red,0.0;green,0.0;blue,0.0}, opacity={1.0}, rotate={0}}, xlabel style={at={(ticklabel cs:0.5)}, anchor=near ticklabel, font={{\fontsize{8 pt}{10.4 pt}\selectfont}}, color={rgb,1:red,0.0;green,0.0;blue,0.0}, draw opacity={1.0}, rotate={0.0}}, xmajorgrids={true}, xmin={1.58}, xmax={16.42}, xtick={{2.0,4.0,6.0,8.0,10.0,12.0,14.0,16.0}}, xticklabels={{$2$,$4$,$6$,$8$,$10$,$12$,$14$,$16$}}, xtick align={inside}, xticklabel style={font={{\fontsize{8 pt}{10.4 pt}\selectfont}}, color={rgb,1:red,0.0;green,0.0;blue,0.0}, draw opacity={1.0}, rotate={0.0}}, x grid style={color={rgb,1:red,0.0;green,0.0;blue,0.0}, draw opacity={0.1}, line width={0.5}, solid}, axis x line*={left}, x axis line style={color={rgb,1:red,0.0;green,0.0;blue,0.0}, draw opacity={1.0}, line width={1}, solid}, scaled y ticks={false}, ylabel={Jumpstart}, y tick style={color={rgb,1:red,0.0;green,0.0;blue,0.0}, opacity={1.0}}, y tick label style={color={rgb,1:red,0.0;green,0.0;blue,0.0}, opacity={1.0}, rotate={0}}, ylabel style={at={(ticklabel cs:0.5)}, anchor=near ticklabel, font={{\fontsize{8 pt}{10.4 pt}\selectfont}}, color={rgb,1:red,0.0;green,0.0;blue,0.0}, draw opacity={1.0}, rotate={0.0}}, ymajorgrids={true}, ymin={-0.11655414791258976}, ymax={0.6707765227885701}, ytick={{0.0,0.2,0.4,0.6000000000000001}}, yticklabels={{$0.0$,$0.2$,$0.4$,$0.6$}}, ytick align={inside}, yticklabel style={font={{\fontsize{8 pt}{10.4 pt}\selectfont}}, color={rgb,1:red,0.0;green,0.0;blue,0.0}, draw opacity={1.0}, rotate={0.0}}, y grid style={color={rgb,1:red,0.0;green,0.0;blue,0.0}, draw opacity={0.1}, line width={0.5}, solid}, axis y line*={left}, y axis line style={color={rgb,1:red,0.0;green,0.0;blue,0.0}, draw opacity={1.0}, line width={1}, solid}]
    \addplot[color={rgb,1:red,0.192;green,0.639;blue,0.329}, name path={43013d14-f9e2-455d-a41b-560572ed3940}, draw opacity={1.0}, line width={1}, solid, mark={square*}, mark size={1.125 pt}, mark repeat={1}, mark options={color={rgb,1:red,0.0;green,0.0;blue,0.0}, draw opacity={1.0}, fill={rgb,1:red,0.192;green,0.639;blue,0.329}, fill opacity={1.0}, line width={0.75}, rotate={0}, solid}]
        table[row sep={\\}]
        {
            \\
            2.0  -0.057088901015636115  \\
            3.0  0.09636811574335614  \\
            4.0  0.021715995082075536  \\
            5.0  0.06001765854688315  \\
            6.0  0.09023070194790225  \\
            7.0  0.36788212399493037  \\
            8.0  0.2064480418441163  \\
            9.0  0.22639727079167704  \\
            10.0  0.08252858165727582  \\
            11.0  0.5070941397654958  \\
            12.0  0.6029412132550375  \\
            13.0  -0.017085409941076588  \\
            14.0  0.3409906267088734  \\
            15.0  0.3127857686742322  \\
            16.0  0.412873282079268  \\
        }
        ;
    \addplot[color={rgb,1:red,0.87;green,0.176;blue,0.149}, name path={85013a87-5d02-4fd7-bb57-34dc7927fbeb}, draw opacity={1.0}, line width={1}, dotted, mark={triangle*}, mark size={1.125 pt}, mark repeat={1}, mark options={color={rgb,1:red,0.0;green,0.0;blue,0.0}, draw opacity={1.0}, fill={rgb,1:red,0.87;green,0.176;blue,0.149}, fill opacity={1.0}, line width={0.75}, rotate={0}, solid}]
        table[row sep={\\}]
        {
            \\
            2.0  -0.07334582574585355  \\
            3.0  0.1970512994778067  \\
            4.0  -0.022249954262080252  \\
            5.0  0.01535748141092623  \\
            6.0  0.05413852045810873  \\
            7.0  0.10310739263181178  \\
            8.0  0.10064128396429531  \\
            9.0  -0.016427381490192582  \\
            10.0  0.014925296278522472  \\
            11.0  -0.031693332252607745  \\
            12.0  0.64849357927816  \\
            13.0  -0.09427120440217958  \\
            14.0  0.01058866116746382  \\
            15.0  -0.057069610200669335  \\
            16.0  0.09957317388792282  \\
        }
        ;
\end{axis}
\end{tikzpicture}
        \caption{\small Jumpstart improvement.}
        \label{fig:gwc_jumpstart}
    \end{subfigure}
    \hfill
    \begin{subfigure}[b]{0.32\textwidth}
        \centering
        \begin{tikzpicture}[/tikz/background rectangle/.style={fill={rgb,1:red,1.0;green,1.0;blue,1.0}, draw opacity={1.0}}, show background rectangle]
\begin{axis}[point meta max={nan}, point meta min={nan}, legend cell align={left}, title={}, title style={at={{(0.5,1)}}, anchor={south}, font={{\fontsize{14 pt}{18.2 pt}\selectfont}}, color={rgb,1:red,0.0;green,0.0;blue,0.0}, draw opacity={1.0}, rotate={0.0}}, legend style={color={rgb,1:red,0.0;green,0.0;blue,0.0}, draw opacity={0.0}, line width={1}, solid, fill={rgb,1:red,1.0;green,1.0;blue,1.0}, fill opacity={1.0}, text opacity={1.0}, font={{\fontsize{8 pt}{10.4 pt}\selectfont}}, at={(0.01, 1.15)}, anchor={south west}, legend columns=-1}, legend style={/tikz/every even column/.append style={column sep=0.5cm}}, axis background/.style={fill={rgb,1:red,1.0;green,1.0;blue,1.0}, opacity={1.0}}, anchor={north west}, xshift={1.0mm}, yshift={-1.0mm}, width={\textwidth}, height={35mm}, scaled x ticks={false}, xlabel={Task Iteration}, x tick style={color={rgb,1:red,0.0;green,0.0;blue,0.0}, opacity={1.0}}, x tick label style={color={rgb,1:red,0.0;green,0.0;blue,0.0}, opacity={1.0}, rotate={0}}, xlabel style={at={(ticklabel cs:0.5)}, anchor=near ticklabel, font={{\fontsize{8 pt}{10.4 pt}\selectfont}}, color={rgb,1:red,0.0;green,0.0;blue,0.0}, draw opacity={1.0}, rotate={0.0}}, xmajorgrids={true}, xmin={1.58}, xmax={16.42}, xtick={{2.0,4.0,6.0,8.0,10.0,12.0,14.0,16.0}}, xticklabels={{$2$,$4$,$6$,$8$,$10$,$12$,$14$,$16$}}, xtick align={inside}, xticklabel style={font={{\fontsize{8 pt}{10.4 pt}\selectfont}}, color={rgb,1:red,0.0;green,0.0;blue,0.0}, draw opacity={1.0}, rotate={0.0}}, x grid style={color={rgb,1:red,0.0;green,0.0;blue,0.0}, draw opacity={0.1}, line width={0.5}, solid}, axis x line*={left}, x axis line style={color={rgb,1:red,0.0;green,0.0;blue,0.0}, draw opacity={1.0}, line width={1}, solid}, scaled y ticks={false}, ylabel={Final Improvement}, y tick style={color={rgb,1:red,0.0;green,0.0;blue,0.0}, opacity={1.0}}, y tick label style={color={rgb,1:red,0.0;green,0.0;blue,0.0}, opacity={1.0}, rotate={0}}, ylabel style={at={(ticklabel cs:0.5)}, anchor=near ticklabel, font={{\fontsize{8 pt}{10.4 pt}\selectfont}}, color={rgb,1:red,0.0;green,0.0;blue,0.0}, draw opacity={1.0}, rotate={0.0}}, ymajorgrids={true}, ymin={-3.5}, ymax={1.8157099648922355}, ytick={{-3.0,-2.0,-1.0,0.0,1.0}}, yticklabels={{$-3$,$-2$,$-1$,$0$,$1$}}, ytick align={inside}, yticklabel style={font={{\fontsize{8 pt}{10.4 pt}\selectfont}}, color={rgb,1:red,0.0;green,0.0;blue,0.0}, draw opacity={1.0}, rotate={0.0}}, y grid style={color={rgb,1:red,0.0;green,0.0;blue,0.0}, draw opacity={0.1}, line width={0.5}, solid}, axis y line*={left}, y axis line style={color={rgb,1:red,0.0;green,0.0;blue,0.0}, draw opacity={1.0}, line width={1}, solid}]
    \addplot[color={rgb,1:red,0.192;green,0.51;blue,0.74}, name path={60d1854e-488d-4066-908f-e5d9a6724d15}, draw opacity={1.0}, line width={1}, dashdotted, mark={*}, mark size={1.125 pt}, mark repeat={1}, mark options={color={rgb,1:red,0.0;green,0.0;blue,0.0}, draw opacity={1.0}, fill={rgb,1:red,0.192;green,0.51;blue,0.74}, fill opacity={1.0}, line width={0.75}, rotate={0}, solid}]
        table[row sep={\\}]
        {
            \\
            2.0  0.321153982711098  \\
            3.0  0.4977817235663589  \\
            4.0  0.37448899863031165  \\
            5.0  0.19617015539007615  \\
            6.0  0.41461570986982954  \\
            7.0  0.2849519888467447  \\
            8.0  -0.25144211285078316  \\
            9.0  0.8657358995084922  \\
            10.0  0.22059755680837717  \\
            11.0  0.46366786208526595  \\
            12.0  0.4866532556208667  \\
            13.0  0.457508894900701  \\
            14.0  1.8157099648922355  \\
            15.0  0.4054797412978537  \\
            16.0  1.784560061207647  \\
        }
        ;
    \addlegendentry {A2T w/ SAVT}
    \addplot[color={rgb,1:red,0.192;green,0.639;blue,0.329}, name path={0e07b849-0290-457e-90ba-9e12e4652316}, draw opacity={1.0}, line width={1}, solid, mark={square*}, mark size={1.125 pt}, mark repeat={1}, mark options={color={rgb,1:red,0.0;green,0.0;blue,0.0}, draw opacity={1.0}, fill={rgb,1:red,0.192;green,0.639;blue,0.329}, fill opacity={1.0}, line width={0.75}, rotate={0}, solid}]
        table[row sep={\\}]
        {
            \\
            2.0  0.16923098481267776  \\
            3.0  -0.09760424241203013  \\
            4.0  0.18479154653138996  \\
            5.0  0.1489361864023487  \\
            6.0  0.19713703686872433  \\
            7.0  0.11419405545845683  \\
            8.0  -0.870819192697296  \\
            9.0  0.16648777270924903  \\
            10.0  -0.05530829544289356  \\
            11.0  -0.4065744748829453  \\
            12.0  -0.16837774537463496  \\
            13.0  -0.1443537468494191  \\
            14.0  1.1963748312646156  \\
            15.0  -0.5164382685861487  \\
            16.0  0.20107728155670188  \\
        }
        ;
    \addlegendentry {A2T}
    \addplot[color={rgb,1:red,0.87;green,0.176;blue,0.149}, name path={1096cd52-7e09-4d6f-b6dc-f4e0ac0efba8}, draw opacity={1.0}, line width={1}, dotted, mark={triangle*}, mark size={1.125 pt}, mark repeat={1}, mark options={color={rgb,1:red,0.0;green,0.0;blue,0.0}, draw opacity={1.0}, fill={rgb,1:red,0.87;green,0.176;blue,0.149}, fill opacity={1.0}, line width={0.75}, rotate={0}, solid}]
        table[row sep={\\}]
        {
            \\
            2.0  -4.64230848123167  \\
            3.0  -2.685891676203674  \\
            4.0  -3.3908419197159096  \\
            5.0  -1.3165956382181547  \\
            6.0  -0.36891006475561766  \\
            7.0  -6.328708725975697  \\
            8.0  -4.657439908097467  \\
            9.0  -6.7561759709610785  \\
            10.0  -2.5664336363886453  \\
            11.0  -10.686849619413254  \\
            12.0  -4.620123177839555  \\
            13.0  -6.487776537112948  \\
            14.0  -15.586104780368824  \\
            15.0  -8.041097202406737  \\
            16.0  -8.091560344757706  \\
        }
        ;
    \addlegendentry {Fine-tune}
\end{axis}
\end{tikzpicture}
        \caption{\small Final improvement.}
        \label{fig:gwc_final}
    \end{subfigure}
    \hfill
    \begin{subfigure}[b]{0.32\textwidth}
        \centering
        \begin{tikzpicture}[/tikz/background rectangle/.style={fill={rgb,1:red,1.0;green,1.0;blue,1.0}, draw opacity={1.0}}, show background rectangle]
\begin{axis}[point meta max={nan}, point meta min={nan}, axis background/.style={fill={rgb,1:red,1.0;green,1.0;blue,1.0}, opacity={1.0}}, anchor={north west}, xshift={1.0mm}, yshift={-1.0mm}, width={\textwidth}, height={35mm}, scaled x ticks={false}, xlabel={Task Iteration}, x tick style={color={rgb,1:red,0.0;green,0.0;blue,0.0}, opacity={1.0}}, x tick label style={color={rgb,1:red,0.0;green,0.0;blue,0.0}, opacity={1.0}, rotate={0}}, xlabel style={at={(ticklabel cs:0.5)}, anchor=near ticklabel, font={{\fontsize{8 pt}{10.4 pt}\selectfont}}, color={rgb,1:red,0.0;green,0.0;blue,0.0}, draw opacity={1.0}, rotate={0.0}}, xmajorgrids={true}, xmin={1.58}, xmax={16.42}, xtick={{2.0,4.0,6.0,8.0,10.0,12.0,14.0,16.0}}, xticklabels={{$2$,$4$,$6$,$8$,$10$,$12$,$14$,$16$}}, xtick align={inside}, xticklabel style={font={{\fontsize{8 pt}{10.4 pt}\selectfont}}, color={rgb,1:red,0.0;green,0.0;blue,0.0}, draw opacity={1.0}, rotate={0.0}}, x grid style={color={rgb,1:red,0.0;green,0.0;blue,0.0}, draw opacity={0.1}, line width={0.5}, solid}, axis x line*={left}, x axis line style={color={rgb,1:red,0.0;green,0.0;blue,0.0}, draw opacity={1.0}, line width={1}, solid}, scaled y ticks={false}, ylabel={Step Ratio}, y tick style={color={rgb,1:red,0.0;green,0.0;blue,0.0}, opacity={1.0}}, y tick label style={color={rgb,1:red,0.0;green,0.0;blue,0.0}, opacity={1.0}, rotate={0}}, ylabel style={at={(ticklabel cs:0.5)}, anchor=near ticklabel, font={{\fontsize{8 pt}{10.4 pt}\selectfont}}, color={rgb,1:red,0.0;green,0.0;blue,0.0}, draw opacity={1.0}, rotate={0.0}}, ymajorgrids={true}, ymin={-0.4}, ymax={2.008514881539175}, ytick={{0.0,0.5,1.0,1.5,2.0}}, yticklabels={{$0.0$,$0.5$,$1.0$,$1.5$,$2.0$}}, ytick align={inside}, yticklabel style={font={{\fontsize{8 pt}{10.4 pt}\selectfont}}, color={rgb,1:red,0.0;green,0.0;blue,0.0}, draw opacity={1.0}, rotate={0.0}}, y grid style={color={rgb,1:red,0.0;green,0.0;blue,0.0}, draw opacity={0.1}, line width={0.5}, solid}, axis y line*={left}, y axis line style={color={rgb,1:red,0.0;green,0.0;blue,0.0}, draw opacity={1.0}, line width={1}, solid}]
    \addplot[color={rgb,1:red,0.192;green,0.51;blue,0.74}, name path={c49ed5f5-a4e6-45e2-b2d5-939843c213d4}, draw opacity={1.0}, line width={1}, dashdotted, mark={*}, mark size={1.125 pt}, mark repeat={1}, mark options={color={rgb,1:red,0.0;green,0.0;blue,0.0}, draw opacity={1.0}, fill={rgb,1:red,0.192;green,0.51;blue,0.74}, fill opacity={1.0}, line width={0.75}, rotate={0}, solid}]
        table[row sep={\\}]
        {
            \\
            2.0  1.0123973146153464  \\
            3.0  0.7283649044876046  \\
            4.0  0.923195145790443  \\
            5.0  0.8725197537972545  \\
            6.0  0.17486737733906593  \\
            7.0  0.7187534401560224  \\
        }
        ;
    \addplot[color={rgb,1:red,0.192;green,0.51;blue,0.74}, name path={c49ed5f5-a4e6-45e2-b2d5-939843c213d4}, draw opacity={1.0}, line width={1}, dashdotted, mark={*}, mark size={1.125 pt}, mark repeat={1}, mark options={color={rgb,1:red,0.0;green,0.0;blue,0.0}, draw opacity={1.0}, fill={rgb,1:red,0.192;green,0.51;blue,0.74}, fill opacity={1.0}, line width={0.75}, rotate={0}, solid}, forget plot]
        table[row sep={\\}]
        {
            \\
            9.0  1.1126704377542918  \\
            10.0  0.8308430202284118  \\
            11.0  0.7437139409762247  \\
            12.0  0.453490202602681  \\
            13.0  1.0482368110116156  \\
            14.0  0.682847010676036  \\
            15.0  0.859893483906584  \\
            16.0  0.8204802750831773  \\
        }
        ;
    \addplot[color={rgb,1:red,0.192;green,0.639;blue,0.329}, name path={5dfa7e46-1b45-4acf-b212-1fea45861bed}, draw opacity={1.0}, line width={1}, solid, mark={square*}, mark size={1.125 pt}, mark repeat={1}, mark options={color={rgb,1:red,0.0;green,0.0;blue,0.0}, draw opacity={1.0}, fill={rgb,1:red,0.192;green,0.639;blue,0.329}, fill opacity={1.0}, line width={0.75}, rotate={0}, solid}]
        table[row sep={\\}]
        {
            \\
            2.0  1.0165288345518757  \\
            3.0  1.1677751856852578  \\
            4.0  1.4457824611709922  \\
            5.0  0.9376738217303455  \\
            6.0  0.5573780166600729  \\
            7.0  0.9687536177124623  \\
        }
        ;
    \addplot[color={rgb,1:red,0.192;green,0.639;blue,0.329}, name path={5dfa7e46-1b45-4acf-b212-1fea45861bed}, draw opacity={1.0}, line width={1}, solid, mark={square*}, mark size={1.125 pt}, mark repeat={1}, mark options={color={rgb,1:red,0.0;green,0.0;blue,0.0}, draw opacity={1.0}, fill={rgb,1:red,0.192;green,0.639;blue,0.329}, fill opacity={1.0}, line width={0.75}, rotate={0}, solid}, forget plot]
        table[row sep={\\}]
        {
            \\
            9.0  1.2392012043200584  \\
            10.0  1.6467205715071511  \\
            11.0  1.7676611746671418  \\
            12.0  0.9624338439768837  \\
            13.0  2.008514881539175  \\
            14.0  0.8386032552937179  \\
        }
        ;
    \addplot[color={rgb,1:red,0.192;green,0.639;blue,0.329}, name path={5dfa7e46-1b45-4acf-b212-1fea45861bed}, draw opacity={1.0}, line width={1}, solid, mark={square*}, mark size={1.125 pt}, mark repeat={1}, mark options={color={rgb,1:red,0.0;green,0.0;blue,0.0}, draw opacity={1.0}, fill={rgb,1:red,0.192;green,0.639;blue,0.329}, fill opacity={1.0}, line width={0.75}, rotate={0}, solid}, forget plot]
        table[row sep={\\}]
        {
            \\
            16.0  1.3927742473375113  \\
        }
        ;
\end{axis}
\end{tikzpicture}
        \caption{\small Step ratio to threshold.}
        \label{fig:gwc_step}
    \end{subfigure}
    \caption{Evaluation metrics for the gridworld scenario with comparable systems.}
    \label{fig:gwc}
    \vskip -0.5 in
\end{figure*}
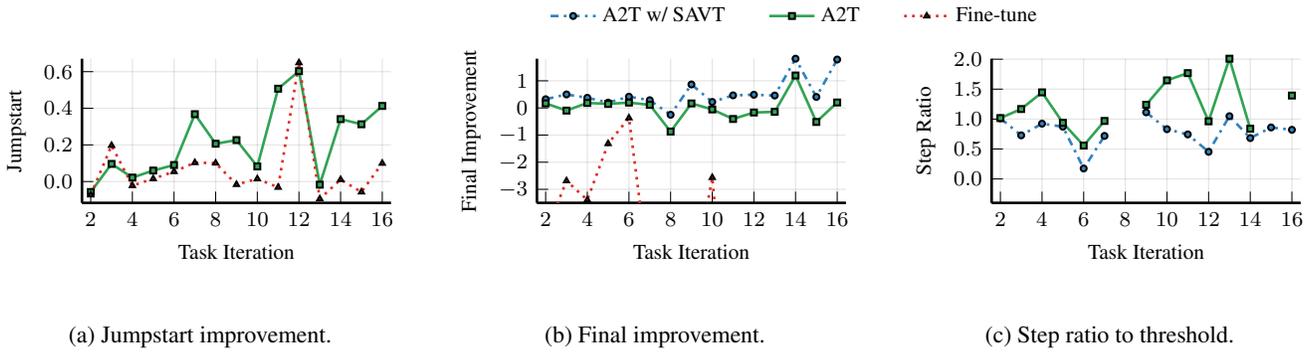

\begin{figure*}
    \centering
    \begin{subfigure}[b]{0.32\textwidth}
        \centering
        \begin{tikzpicture}[/tikz/background rectangle/.style={fill={rgb,1:red,1.0;green,1.0;blue,1.0}, draw opacity={1.0}}, show background rectangle]
\begin{axis}[point meta max={nan}, point meta min={nan}, axis background/.style={fill={rgb,1:red,1.0;green,1.0;blue,1.0}, opacity={1.0}}, anchor={north west}, xshift={1.0mm}, yshift={-1.0mm}, width=\textwidth, height={35mm}, scaled x ticks={false}, xlabel={Task Iteration}, x tick style={color={rgb,1:red,0.0;green,0.0;blue,0.0}, opacity={1.0}}, x tick label style={color={rgb,1:red,0.0;green,0.0;blue,0.0}, opacity={1.0}, rotate={0}}, xlabel style={at={(ticklabel cs:0.5)}, anchor=near ticklabel, font={{\fontsize{8 pt}{10.4 pt}\selectfont}}, color={rgb,1:red,0.0;green,0.0;blue,0.0}, draw opacity={1.0}, rotate={0.0}}, xmajorgrids={true}, xmin={1.76}, xmax={10.24}, xtick={{2.0,4.0,6.0,8.0,10.0}}, xticklabels={{$2$,$4$,$6$,$8$,$10$}}, xtick align={inside}, xticklabel style={font={{\fontsize{8 pt}{10.4 pt}\selectfont}}, color={rgb,1:red,0.0;green,0.0;blue,0.0}, draw opacity={1.0}, rotate={0.0}}, x grid style={color={rgb,1:red,0.0;green,0.0;blue,0.0}, draw opacity={0.1}, line width={0.5}, solid}, axis x line*={left}, x axis line style={color={rgb,1:red,0.0;green,0.0;blue,0.0}, draw opacity={1.0}, line width={1}, solid}, scaled y ticks={false}, ylabel={Jumpstart}, y tick style={color={rgb,1:red,0.0;green,0.0;blue,0.0}, opacity={1.0}}, y tick label style={color={rgb,1:red,0.0;green,0.0;blue,0.0}, opacity={1.0}, rotate={0}}, ylabel style={at={(ticklabel cs:0.5)}, anchor=near ticklabel, font={{\fontsize{8 pt}{10.4 pt}\selectfont}}, color={rgb,1:red,0.0;green,0.0;blue,0.0}, draw opacity={1.0}, rotate={0.0}}, ymajorgrids={true}, ymin={0}, ymax={6.745248638712139}, ytick={{0.0,1.0,2.0,3.0,4.0,5.0,6.0}}, yticklabels={{$0$,$1$,$2$,$3$,$4$,$5$,$6$}}, ytick align={inside}, yticklabel style={font={{\fontsize{8 pt}{10.4 pt}\selectfont}}, color={rgb,1:red,0.0;green,0.0;blue,0.0}, draw opacity={1.0}, rotate={0.0}}, y grid style={color={rgb,1:red,0.0;green,0.0;blue,0.0}, draw opacity={0.1}, line width={0.5}, solid}, axis y line*={left}, y axis line style={color={rgb,1:red,0.0;green,0.0;blue,0.0}, draw opacity={1.0}, line width={1}, solid}]
    \addplot[color={rgb,1:red,0.192;green,0.639;blue,0.329}, name path={a77b43ad-5f52-496d-b360-e29d6a2ca221}, draw opacity={1.0}, line width={1}, solid, mark={square*}, mark size={1.125 pt}, mark repeat={1}, mark options={color={rgb,1:red,0.0;green,0.0;blue,0.0}, draw opacity={1.0}, fill={rgb,1:red,0.192;green,0.639;blue,0.329}, fill opacity={1.0}, line width={0.75}, rotate={0}, solid}]
        table[row sep={\\}]
        {
            \\
            2.0  2.147920878577009  \\
            3.0  6.745248638712139  \\
            4.0  6.143082103103621  \\
            5.0  2.5976881443105806  \\
            6.0  2.8629947266135054  \\
            7.0  2.0746141979235415  \\
            8.0  1.4274282088044445  \\
            9.0  1.1168549616442531  \\
            10.0  1.5496410283538251  \\
        }
        ;
    \addplot[color={rgb,1:red,0.87;green,0.176;blue,0.149}, name path={f70acd8b-d5ac-4879-9a27-97ecc6644e92}, draw opacity={1.0}, line width={1}, dotted, mark={triangle*}, mark size={1.125 pt}, mark repeat={1}, mark options={color={rgb,1:red,0.0;green,0.0;blue,0.0}, draw opacity={1.0}, fill={rgb,1:red,0.87;green,0.176;blue,0.149}, fill opacity={1.0}, line width={0.75}, rotate={0}, solid}]
        table[row sep={\\}]
        {
            \\
            2.0  2.3398142103538087  \\
            3.0  5.8636884845601625  \\
            4.0  4.821441629995734  \\
            5.0  1.5910003249041889  \\
            6.0  2.469657917197317  \\
            7.0  1.878860631122166  \\
            8.0  1.5822434830609673  \\
            9.0  1.6103306484936237  \\
            10.0  1.9826832837651158  \\
        }
        ;
\end{axis}
\end{tikzpicture}
        \caption{Jumpstart improvement.}
        \label{fig:adl_jumpstart}
    \end{subfigure}
    \hfill
    \begin{subfigure}[b]{0.32\textwidth}
        \centering
        \begin{tikzpicture}[/tikz/background rectangle/.style={fill={rgb,1:red,1.0;green,1.0;blue,1.0}, draw opacity={1.0}}, show background rectangle]
\begin{axis}[point meta max={nan}, point meta min={nan}, legend cell align={left}, title={}, title style={at={{(0.5,1)}}, anchor={south}, font={{\fontsize{14 pt}{18.2 pt}\selectfont}}, color={rgb,1:red,0.0;green,0.0;blue,0.0}, draw opacity={1.0}, rotate={0.0}}, legend style={color={rgb,1:red,0.0;green,0.0;blue,0.0}, draw opacity={0.0}, line width={1}, solid, fill={rgb,1:red,1.0;green,1.0;blue,1.0}, fill opacity={1.0}, text opacity={1.0}, font={{\fontsize{8 pt}{10.4 pt}\selectfont}}, at={(0.01, 1.15)}, anchor={south west}, legend columns=-1}, legend style={/tikz/every even column/.append style={column sep=0.5cm}}, axis background/.style={fill={rgb,1:red,1.0;green,1.0;blue,1.0}, opacity={1.0}}, anchor={north west}, xshift={1.0mm}, yshift={-1.0mm}, width=\textwidth, height={35mm}, scaled x ticks={false}, xlabel={Task Iteration}, x tick style={color={rgb,1:red,0.0;green,0.0;blue,0.0}, opacity={1.0}}, x tick label style={color={rgb,1:red,0.0;green,0.0;blue,0.0}, opacity={1.0}, rotate={0}}, xlabel style={at={(ticklabel cs:0.5)}, anchor=near ticklabel, font={{\fontsize{8 pt}{10.4 pt}\selectfont}}, color={rgb,1:red,0.0;green,0.0;blue,0.0}, draw opacity={1.0}, rotate={0.0}}, xmajorgrids={true}, xmin={1.76}, xmax={10.24}, xtick={{2.0,4.0,6.0,8.0,10.0}}, xticklabels={{$2$,$4$,$6$,$8$,$10$}}, xtick align={inside}, xticklabel style={font={{\fontsize{8 pt}{10.4 pt}\selectfont}}, color={rgb,1:red,0.0;green,0.0;blue,0.0}, draw opacity={1.0}, rotate={0.0}}, x grid style={color={rgb,1:red,0.0;green,0.0;blue,0.0}, draw opacity={0.1}, line width={0.5}, solid}, axis x line*={left}, x axis line style={color={rgb,1:red,0.0;green,0.0;blue,0.0}, draw opacity={1.0}, line width={1}, solid}, scaled y ticks={false}, ylabel={Final Improvement}, y tick style={color={rgb,1:red,0.0;green,0.0;blue,0.0}, opacity={1.0}}, y tick label style={color={rgb,1:red,0.0;green,0.0;blue,0.0}, opacity={1.0}, rotate={0}}, ylabel style={at={(ticklabel cs:0.5)}, anchor=near ticklabel, font={{\fontsize{8 pt}{10.4 pt}\selectfont}}, color={rgb,1:red,0.0;green,0.0;blue,0.0}, draw opacity={1.0}, rotate={0.0}}, ymajorgrids={true}, ymin={-0.5179410983433456}, ymax={7.679877224377678}, ytick={{0.0,2.0,4.0,6.0}}, yticklabels={{$0$,$2$,$4$,$6$}}, ytick align={inside}, yticklabel style={font={{\fontsize{8 pt}{10.4 pt}\selectfont}}, color={rgb,1:red,0.0;green,0.0;blue,0.0}, draw opacity={1.0}, rotate={0.0}}, y grid style={color={rgb,1:red,0.0;green,0.0;blue,0.0}, draw opacity={0.1}, line width={0.5}, solid}, axis y line*={left}, y axis line style={color={rgb,1:red,0.0;green,0.0;blue,0.0}, draw opacity={1.0}, line width={1}, solid}]
    \addplot[color={rgb,1:red,0.192;green,0.51;blue,0.74}, name path={eb10e79f-566c-43a9-ad5f-6cad784ae6e7}, draw opacity={1.0}, line width={1}, dashdotted, mark={*}, mark size={1.125 pt}, mark repeat={1}, mark options={color={rgb,1:red,0.0;green,0.0;blue,0.0}, draw opacity={1.0}, fill={rgb,1:red,0.192;green,0.51;blue,0.74}, fill opacity={1.0}, line width={0.75}, rotate={0}, solid}]
        table[row sep={\\}]
        {
            \\
            2.0  0.4095767033659284  \\
            3.0  0.04549736844681955  \\
            4.0  1.568791249855136  \\
            5.0  2.4583000048325587  \\
            6.0  3.1078000701185298  \\
            7.0  7.290087369653672  \\
            8.0  1.3900458559109918  \\
            9.0  0.646451093069577  \\
            10.0  0.6728101147485993  \\
        }
        ;
    \addlegendentry {A2T w/ SAVT}
    \addplot[color={rgb,1:red,0.192;green,0.639;blue,0.329}, name path={fcc77504-501c-4331-8e81-8b7b20b24918}, draw opacity={1.0}, line width={1}, solid, mark={square*}, mark size={1.125 pt}, mark repeat={1}, mark options={color={rgb,1:red,0.0;green,0.0;blue,0.0}, draw opacity={1.0}, fill={rgb,1:red,0.192;green,0.639;blue,0.329}, fill opacity={1.0}, line width={0.75}, rotate={0}, solid}]
        table[row sep={\\}]
        {
            \\
            2.0  0.1746796283304938  \\
            3.0  0.05237274014865093  \\
            4.0  1.6027153512660341  \\
            5.0  2.4335111710294957  \\
            6.0  3.1065781528538667  \\
            7.0  7.447863498262931  \\
            8.0  2.995524033786952  \\
            9.0  0.13042295105087504  \\
            10.0  0.5270413109832387  \\
        }
        ;
    \addlegendentry {A2T}
    \addplot[color={rgb,1:red,0.87;green,0.176;blue,0.149}, name path={dd378f72-8cab-40be-afa2-3280ffa03a08}, draw opacity={1.0}, line width={1}, dotted, mark={triangle*}, mark size={1.125 pt}, mark repeat={1}, mark options={color={rgb,1:red,0.0;green,0.0;blue,0.0}, draw opacity={1.0}, fill={rgb,1:red,0.87;green,0.176;blue,0.149}, fill opacity={1.0}, line width={0.75}, rotate={0}, solid}]
        table[row sep={\\}]
        {
            \\
            2.0  -0.2436203129697768  \\
            3.0  -0.08741652273974422  \\
            4.0  1.332331855182098  \\
            5.0  0.7624689041107812  \\
            6.0  -0.021072706173591912  \\
            7.0  0.712618345244703  \\
            8.0  0.3590374304676264  \\
            9.0  -0.2859273722285996  \\
            10.0  -0.0707768684512639  \\
        }
        ;
    \addlegendentry {Fine-tune}
\end{axis}
\end{tikzpicture}
        \caption{Final improvement.}
        \label{fig:adl_final}
    \end{subfigure}
    \hfill
    \begin{subfigure}[b]{0.32\textwidth}
        \centering
        \begin{tikzpicture}[/tikz/background rectangle/.style={fill={rgb,1:red,1.0;green,1.0;blue,1.0}, draw opacity={1.0}}, show background rectangle]
\begin{axis}[point meta max={nan}, point meta min={nan}, axis background/.style={fill={rgb,1:red,1.0;green,1.0;blue,1.0}, opacity={1.0}}, anchor={north west}, xshift={1.0mm}, yshift={-1.0mm}, width=\textwidth, height={35mm}, scaled x ticks={false}, xlabel={Task Iteration}, x tick style={color={rgb,1:red,0.0;green,0.0;blue,0.0}, opacity={1.0}}, x tick label style={color={rgb,1:red,0.0;green,0.0;blue,0.0}, opacity={1.0}, rotate={0}}, xlabel style={at={(ticklabel cs:0.5)}, anchor=near ticklabel, font={{\fontsize{8 pt}{10.4 pt}\selectfont}}, color={rgb,1:red,0.0;green,0.0;blue,0.0}, draw opacity={1.0}, rotate={0.0}}, xmajorgrids={true}, xmin={1.76}, xmax={10.24}, xtick={{2.0,4.0,6.0,8.0,10.0}}, xticklabels={{$2$,$4$,$6$,$8$,$10$}}, xtick align={inside}, xticklabel style={font={{\fontsize{8 pt}{10.4 pt}\selectfont}}, color={rgb,1:red,0.0;green,0.0;blue,0.0}, draw opacity={1.0}, rotate={0.0}}, x grid style={color={rgb,1:red,0.0;green,0.0;blue,0.0}, draw opacity={0.1}, line width={0.5}, solid}, axis x line*={left}, x axis line style={color={rgb,1:red,0.0;green,0.0;blue,0.0}, draw opacity={1.0}, line width={1}, solid}, scaled y ticks={false}, ylabel={Step Ratio}, y tick style={color={rgb,1:red,0.0;green,0.0;blue,0.0}, opacity={1.0}}, y tick label style={color={rgb,1:red,0.0;green,0.0;blue,0.0}, opacity={1.0}, rotate={0}}, ylabel style={at={(ticklabel cs:0.5)}, anchor=near ticklabel, font={{\fontsize{8 pt}{10.4 pt}\selectfont}}, color={rgb,1:red,0.0;green,0.0;blue,0.0}, draw opacity={1.0}, rotate={0.0}}, ymajorgrids={true}, ymin={-0.034800548030131}, ymax={1.5272557672401859}, ytick={{0.0,0.5,1.0,1.5}}, yticklabels={{$0.0$,$0.5$,$1.0$,$1.5$}}, ytick align={inside}, yticklabel style={font={{\fontsize{8 pt}{10.4 pt}\selectfont}}, color={rgb,1:red,0.0;green,0.0;blue,0.0}, draw opacity={1.0}, rotate={0.0}}, y grid style={color={rgb,1:red,0.0;green,0.0;blue,0.0}, draw opacity={0.1}, line width={0.5}, solid}, axis y line*={left}, y axis line style={color={rgb,1:red,0.0;green,0.0;blue,0.0}, draw opacity={1.0}, line width={1}, solid}]
    \addplot[color={rgb,1:red,0.192;green,0.51;blue,0.74}, name path={400c0095-1610-4b3e-a386-db8a28fd2330}, draw opacity={1.0}, line width={1}, dashdotted, mark={*}, mark size={1.125 pt}, mark repeat={1}, mark options={color={rgb,1:red,0.0;green,0.0;blue,0.0}, draw opacity={1.0}, fill={rgb,1:red,0.192;green,0.51;blue,0.74}, fill opacity={1.0}, line width={0.75}, rotate={0}, solid}]
        table[row sep={\\}]
        {
            \\
            2.0  0.5951670422721834  \\
            3.0  0.3638993244607355  \\
            4.0  0.009408592968085506  \\
            5.0  0.03575348954358687  \\
            6.0  0.022288517824295986  \\
            7.0  0.015779583679361147  \\
            8.0  0.08496346560361387  \\
            9.0  0.2834616655735326  \\
            10.0  0.22513013406229934  \\
        }
        ;
    \addplot[color={rgb,1:red,0.192;green,0.639;blue,0.329}, name path={445b3ca8-84b4-4dc6-a9a4-1daae1194a0f}, draw opacity={1.0}, line width={1}, solid, mark={square*}, mark size={1.125 pt}, mark repeat={1}, mark options={color={rgb,1:red,0.0;green,0.0;blue,0.0}, draw opacity={1.0}, fill={rgb,1:red,0.192;green,0.639;blue,0.329}, fill opacity={1.0}, line width={0.75}, rotate={0}, solid}]
        table[row sep={\\}]
        {
            \\
            2.0  0.2986797743658294  \\
            3.0  0.30581176502696106  \\
            4.0  0.024786422591212328  \\
            5.0  0.030077929694108094  \\
            6.0  0.017515870404021614  \\
            7.0  0.015753117469200842  \\
            8.0  0.1235625138753487  \\
            9.0  0.5215377647118179  \\
            10.0  0.3623457853065998  \\
        }
        ;
    \addplot[color={rgb,1:red,0.87;green,0.176;blue,0.149}, name path={72103ef0-d113-40b1-9cb1-851945b7bb7f}, draw opacity={1.0}, line width={1}, dotted, mark={triangle*}, mark size={1.125 pt}, mark repeat={1}, mark options={color={rgb,1:red,0.0;green,0.0;blue,0.0}, draw opacity={1.0}, fill={rgb,1:red,0.87;green,0.176;blue,0.149}, fill opacity={1.0}, line width={0.75}, rotate={0}, solid}]
        table[row sep={\\}]
        {
            \\
            4.0  0.032461887406020816  \\
            5.0  1.4830466262419693  \\
            6.0  0.03340058282896212  \\
            7.0  0.03005130868256213  \\
            8.0  0.5540728371347767  \\
        }
        ;
    \addplot[color={rgb,1:red,0.87;green,0.176;blue,0.149}, name path={72103ef0-d113-40b1-9cb1-851945b7bb7f}, draw opacity={1.0}, line width={1}, dotted, mark={triangle*}, mark size={1.125 pt}, mark repeat={1}, mark options={color={rgb,1:red,0.0;green,0.0;blue,0.0}, draw opacity={1.0}, fill={rgb,1:red,0.87;green,0.176;blue,0.149}, fill opacity={1.0}, line width={0.75}, rotate={0}, solid}, forget plot]
        table[row sep={\\}]
        {
            \\
            10.0  0.419734368665942  \\
        }
        ;
\end{axis}
\end{tikzpicture}
        \caption{Step ratio to threshold.}
        \label{fig:adl_step}
    \end{subfigure}
    \caption{Evaluation metrics for the autonomous driving scenario with a learning system.}
    \label{fig:adl}
    \vskip -0.5 in
\end{figure*}
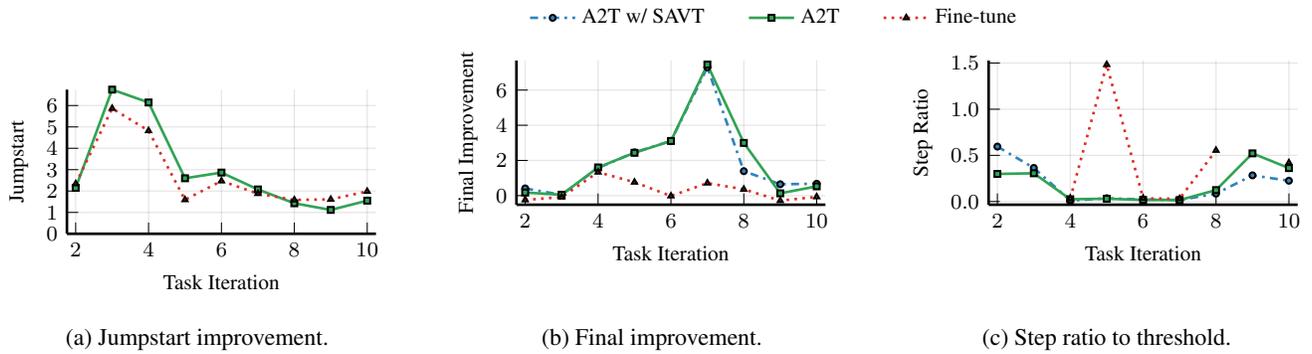

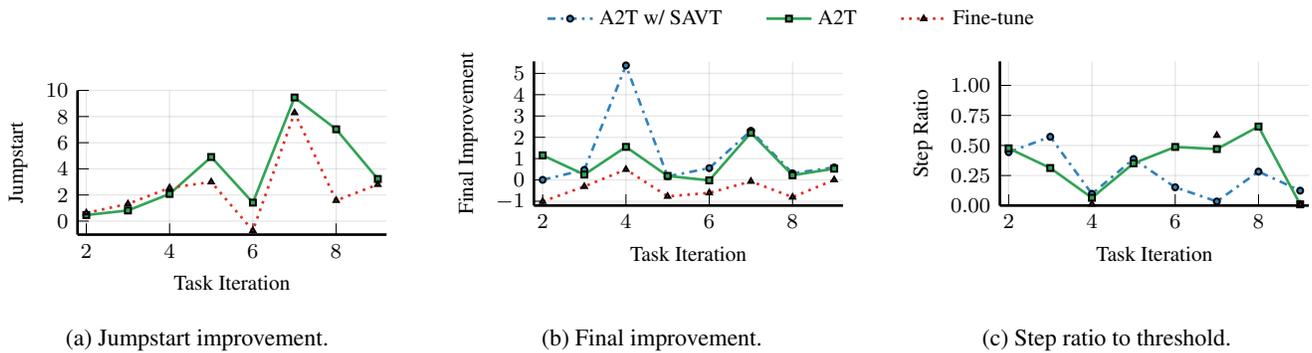
\begin{figure*}
    \centering
    \begin{subfigure}[b]{0.32\textwidth}
        \centering
        \begin{tikzpicture}[/tikz/background rectangle/.style={fill={rgb,1:red,1.0;green,1.0;blue,1.0}, draw opacity={1.0}}, show background rectangle]
\begin{axis}[point meta max={nan}, point meta min={nan}, axis background/.style={fill={rgb,1:red,1.0;green,1.0;blue,1.0}, opacity={1.0}}, anchor={north west}, xshift={1.0mm}, yshift={-1.0mm}, width=\textwidth, height={35mm}, scaled x ticks={false}, xlabel={Task Iteration}, x tick style={color={rgb,1:red,0.0;green,0.0;blue,0.0}, opacity={1.0}}, x tick label style={color={rgb,1:red,0.0;green,0.0;blue,0.0}, opacity={1.0}, rotate={0}}, xlabel style={at={(ticklabel cs:0.5)}, anchor=near ticklabel, font={{\fontsize{8 pt}{10.4 pt}\selectfont}}, color={rgb,1:red,0.0;green,0.0;blue,0.0}, draw opacity={1.0}, rotate={0.0}}, xmajorgrids={true}, xmin={1.79}, xmax={9.21}, xtick={{2.0,4.0,6.0,8.0}}, xticklabels={{$2$,$4$,$6$,$8$}}, xtick align={inside}, xticklabel style={font={{\fontsize{8 pt}{10.4 pt}\selectfont}}, color={rgb,1:red,0.0;green,0.0;blue,0.0}, draw opacity={1.0}, rotate={0.0}}, x grid style={color={rgb,1:red,0.0;green,0.0;blue,0.0}, draw opacity={0.1}, line width={0.5}, solid}, axis x line*={left}, x axis line style={color={rgb,1:red,0.0;green,0.0;blue,0.0}, draw opacity={1.0}, line width={1}, solid}, scaled y ticks={false}, ylabel={Jumpstart}, y tick style={color={rgb,1:red,0.0;green,0.0;blue,0.0}, opacity={1.0}}, y tick label style={color={rgb,1:red,0.0;green,0.0;blue,0.0}, opacity={1.0}, rotate={0}}, ylabel style={at={(ticklabel cs:0.5)}, anchor=near ticklabel, font={{\fontsize{8 pt}{10.4 pt}\selectfont}}, color={rgb,1:red,0.0;green,0.0;blue,0.0}, draw opacity={1.0}, rotate={0.0}}, ymajorgrids={true}, ymin={-1.033265075863943}, ymax={10.0}, ytick={{0.0,2.0,4.0,6.0,8.0,10.0}}, yticklabels={{$0$,$2$,$4$,$6$,$8$,$10$}}, ytick align={inside}, yticklabel style={font={{\fontsize{8 pt}{10.4 pt}\selectfont}}, color={rgb,1:red,0.0;green,0.0;blue,0.0}, draw opacity={1.0}, rotate={0.0}}, y grid style={color={rgb,1:red,0.0;green,0.0;blue,0.0}, draw opacity={0.1}, line width={0.5}, solid}, axis y line*={left}, y axis line style={color={rgb,1:red,0.0;green,0.0;blue,0.0}, draw opacity={1.0}, line width={1}, solid}]
    \addplot[color={rgb,1:red,0.192;green,0.639;blue,0.329}, name path={a77b43ad-5f52-496d-b360-e29d6a2ca221}, draw opacity={1.0}, line width={1}, solid, mark={square*}, mark size={1.125 pt}, mark repeat={1}, mark options={color={rgb,1:red,0.0;green,0.0;blue,0.0}, draw opacity={1.0}, fill={rgb,1:red,0.192;green,0.639;blue,0.329}, fill opacity={1.0}, line width={0.75}, rotate={0}, solid}]
        table[row sep={\\}]
        {
            \\
            2.0  0.4652414314767877  \\
            3.0  0.8142250027949006  \\
            4.0  2.080367113742092  \\
            5.0  4.90565888557582  \\
            6.0  1.4138021036411665  \\
            7.0  9.458310150701209  \\
            8.0  7.02674691497914  \\
            9.0  3.2232678662779106  \\
        }
        ;
    \addplot[color={rgb,1:red,0.87;green,0.176;blue,0.149}, name path={f70acd8b-d5ac-4879-9a27-97ecc6644e92}, draw opacity={1.0}, line width={1}, dotted, mark={triangle*}, mark size={1.125 pt}, mark repeat={1}, mark options={color={rgb,1:red,0.0;green,0.0;blue,0.0}, draw opacity={1.0}, fill={rgb,1:red,0.87;green,0.176;blue,0.149}, fill opacity={1.0}, line width={0.75}, rotate={0}, solid}]
        table[row sep={\\}]
        {
            \\
            2.0  0.6263840026563283  \\
            3.0  1.3157619389390085  \\
            4.0  2.544118852868548  \\
            5.0  2.9967741691526144  \\
            6.0  -0.727685214896026  \\
            7.0  8.290358788230726  \\
            8.0  1.5781337943043752  \\
            9.0  2.80105117467212  \\
        }
        ;
\end{axis}
\end{tikzpicture}
        \caption{Jumpstart improvement.}
        \label{fig:adc_jumpstart}
    \end{subfigure}
    \hfill
    \begin{subfigure}[b]{0.32\textwidth}
        \centering
        \begin{tikzpicture}[/tikz/background rectangle/.style={fill={rgb,1:red,1.0;green,1.0;blue,1.0}, draw opacity={1.0}}, show background rectangle]
\begin{axis}[point meta max={nan}, point meta min={nan}, legend cell align={left}, title={}, title style={at={{(0.5,1)}}, anchor={south}, font={{\fontsize{14 pt}{18.2 pt}\selectfont}}, color={rgb,1:red,0.0;green,0.0;blue,0.0}, draw opacity={1.0}, rotate={0.0}}, legend style={color={rgb,1:red,0.0;green,0.0;blue,0.0}, draw opacity={0.0}, line width={1}, solid, fill={rgb,1:red,1.0;green,1.0;blue,1.0}, fill opacity={1.0}, text opacity={1.0}, font={{\fontsize{8 pt}{10.4 pt}\selectfont}}, at={(0.01, 1.15)}, anchor={south west}, legend columns=-1}, legend style={/tikz/every even column/.append style={column sep=0.5cm}}, axis background/.style={fill={rgb,1:red,1.0;green,1.0;blue,1.0}, opacity={1.0}}, anchor={north west}, xshift={1.0mm}, yshift={-1.0mm}, width=\textwidth, height={35mm}, scaled x ticks={false}, xlabel={Task Iteration}, x tick style={color={rgb,1:red,0.0;green,0.0;blue,0.0}, opacity={1.0}}, x tick label style={color={rgb,1:red,0.0;green,0.0;blue,0.0}, opacity={1.0}, rotate={0}}, xlabel style={at={(ticklabel cs:0.5)}, anchor=near ticklabel, font={{\fontsize{8 pt}{10.4 pt}\selectfont}}, color={rgb,1:red,0.0;green,0.0;blue,0.0}, draw opacity={1.0}, rotate={0.0}}, xmajorgrids={true}, xmin={1.79}, xmax={9.21}, xtick={{2.0,4.0,6.0,8.0}}, xticklabels={{$2$,$4$,$6$,$8$}}, xtick align={inside}, xticklabel style={font={{\fontsize{8 pt}{10.4 pt}\selectfont}}, color={rgb,1:red,0.0;green,0.0;blue,0.0}, draw opacity={1.0}, rotate={0.0}}, x grid style={color={rgb,1:red,0.0;green,0.0;blue,0.0}, draw opacity={0.1}, line width={0.5}, solid}, axis x line*={left}, x axis line style={color={rgb,1:red,0.0;green,0.0;blue,0.0}, draw opacity={1.0}, line width={1}, solid}, scaled y ticks={false}, ylabel={Final Improvement}, y tick style={color={rgb,1:red,0.0;green,0.0;blue,0.0}, opacity={1.0}}, y tick label style={color={rgb,1:red,0.0;green,0.0;blue,0.0}, opacity={1.0}, rotate={0}}, ylabel style={at={(ticklabel cs:0.5)}, anchor=near ticklabel, font={{\fontsize{8 pt}{10.4 pt}\selectfont}}, color={rgb,1:red,0.0;green,0.0;blue,0.0}, draw opacity={1.0}, rotate={0.0}}, ymajorgrids={true}, ymin={-1.2030535485183285}, ymax={5.564286596733486}, ytick={{-1.0,0.0,1.0,2.0,3.0,4.0,5.0}}, yticklabels={{$-1$,$0$,$1$,$2$,$3$,$4$,$5$}}, ytick align={inside}, yticklabel style={font={{\fontsize{8 pt}{10.4 pt}\selectfont}}, color={rgb,1:red,0.0;green,0.0;blue,0.0}, draw opacity={1.0}, rotate={0.0}}, y grid style={color={rgb,1:red,0.0;green,0.0;blue,0.0}, draw opacity={0.1}, line width={0.5}, solid}, axis y line*={left}, y axis line style={color={rgb,1:red,0.0;green,0.0;blue,0.0}, draw opacity={1.0}, line width={1}, solid}]
    \addplot[color={rgb,1:red,0.192;green,0.51;blue,0.74}, name path={602a1b61-b80b-498d-b71d-980cc9010589}, draw opacity={1.0}, line width={1}, dashdotted, mark={*}, mark size={1.125 pt}, mark repeat={1}, mark options={color={rgb,1:red,0.0;green,0.0;blue,0.0}, draw opacity={1.0}, fill={rgb,1:red,0.192;green,0.51;blue,0.74}, fill opacity={1.0}, line width={0.75}, rotate={0}, solid}]
        table[row sep={\\}]
        {
            \\
            2.0  0.003867066233056613  \\
            3.0  0.47079997936676166  \\
            4.0  5.372758102056547  \\
            5.0  0.16515398009070958  \\
            6.0  0.552755999651476  \\
            7.0  2.305185995297521  \\
            8.0  0.30969537246011314  \\
            9.0  0.5877394937122495  \\
        }
        ;
    \addlegendentry {A2T w/ SAVT}
    \addplot[color={rgb,1:red,0.192;green,0.639;blue,0.329}, name path={54b65b61-bcf4-49e4-a711-ff9f72c07796}, draw opacity={1.0}, line width={1}, solid, mark={square*}, mark size={1.125 pt}, mark repeat={1}, mark options={color={rgb,1:red,0.0;green,0.0;blue,0.0}, draw opacity={1.0}, fill={rgb,1:red,0.192;green,0.639;blue,0.329}, fill opacity={1.0}, line width={0.75}, rotate={0}, solid}]
        table[row sep={\\}]
        {
            \\
            2.0  1.1546379426501103  \\
            3.0  0.24989630202381224  \\
            4.0  1.5500224670617402  \\
            5.0  0.18702189020920928  \\
            6.0  -0.021095137622192513  \\
            7.0  2.2123948587999074  \\
            8.0  0.21879484791006365  \\
            9.0  0.5274164315677192  \\
        }
        ;
    \addlegendentry {A2T}
    \addplot[color={rgb,1:red,0.87;green,0.176;blue,0.149}, name path={1b2b1bf1-0d82-436d-a13e-e6a7af374018}, draw opacity={1.0}, line width={1}, dotted, mark={triangle*}, mark size={1.125 pt}, mark repeat={1}, mark options={color={rgb,1:red,0.0;green,0.0;blue,0.0}, draw opacity={1.0}, fill={rgb,1:red,0.87;green,0.176;blue,0.149}, fill opacity={1.0}, line width={0.75}, rotate={0}, solid}]
        table[row sep={\\}]
        {
            \\
            2.0  -1.0115250538413902  \\
            3.0  -0.30971115593463516  \\
            4.0  0.4908587746930845  \\
            5.0  -0.7707491822611804  \\
            6.0  -0.6096691154522278  \\
            7.0  -0.06733836057691077  \\
            8.0  -0.8084954269995686  \\
            9.0  0.003754130064385702  \\
        }
        ;
    \addlegendentry {Fine-tune}
\end{axis}
\end{tikzpicture}
        \caption{Final improvement.}
        \label{fig:adc_final}
    \end{subfigure}
    \hfill
    \begin{subfigure}[b]{0.32\textwidth}
        \centering
        \begin{tikzpicture}[/tikz/background rectangle/.style={fill={rgb,1:red,1.0;green,1.0;blue,1.0}, draw opacity={1.0}}, show background rectangle]
\begin{axis}[point meta max={nan}, point meta min={nan}, axis background/.style={fill={rgb,1:red,1.0;green,1.0;blue,1.0}, opacity={1.0}}, anchor={north west}, xshift={1.0mm}, yshift={-1.0mm}, width=\textwidth, height={35mm}, scaled x ticks={false}, xlabel={Task Iteration}, x tick style={color={rgb,1:red,0.0;green,0.0;blue,0.0}, opacity={1.0}}, x tick label style={color={rgb,1:red,0.0;green,0.0;blue,0.0}, opacity={1.0}, rotate={0}}, xlabel style={at={(ticklabel cs:0.5)}, anchor=near ticklabel, font={{\fontsize{8 pt}{10.4 pt}\selectfont}}, color={rgb,1:red,0.0;green,0.0;blue,0.0}, draw opacity={1.0}, rotate={0.0}}, xmajorgrids={true}, xmin={1.79}, xmax={9.21}, xtick={{2.0,4.0,6.0,8.0}}, xticklabels={{$2$,$4$,$6$,$8$}}, xtick align={inside}, xticklabel style={font={{\fontsize{8 pt}{10.4 pt}\selectfont}}, color={rgb,1:red,0.0;green,0.0;blue,0.0}, draw opacity={1.0}, rotate={0.0}}, x grid style={color={rgb,1:red,0.0;green,0.0;blue,0.0}, draw opacity={0.1}, line width={0.5}, solid}, axis x line*={left}, x axis line style={color={rgb,1:red,0.0;green,0.0;blue,0.0}, draw opacity={1.0}, line width={1}, solid}, scaled y ticks={false}, ylabel={Step Ratio}, y tick style={color={rgb,1:red,0.0;green,0.0;blue,0.0}, opacity={1.0}}, y tick label style={color={rgb,1:red,0.0;green,0.0;blue,0.0}, opacity={1.0}, rotate={0}}, ylabel style={at={(ticklabel cs:0.5)}, anchor=near ticklabel, font={{\fontsize{8 pt}{10.4 pt}\selectfont}}, color={rgb,1:red,0.0;green,0.0;blue,0.0}, draw opacity={1.0}, rotate={0.0}}, ymajorgrids={true}, ymin={0}, ymax={1.2}, ytick={{0.0,0.25,0.5,0.75,1.0}}, yticklabels={{$0.00$,$0.25$,$0.50$,$0.75$,$1.00$}}, ytick align={inside}, yticklabel style={font={{\fontsize{8 pt}{10.4 pt}\selectfont}}, color={rgb,1:red,0.0;green,0.0;blue,0.0}, draw opacity={1.0}, rotate={0.0}}, y grid style={color={rgb,1:red,0.0;green,0.0;blue,0.0}, draw opacity={0.1}, line width={0.5}, solid}, axis y line*={left}, y axis line style={color={rgb,1:red,0.0;green,0.0;blue,0.0}, draw opacity={1.0}, line width={1}, solid}]
    \addplot[color={rgb,1:red,0.192;green,0.51;blue,0.74}, name path={271aeed6-16ca-47e7-a926-b9e2bfceb8a8}, draw opacity={1.0}, line width={1}, dashdotted, mark={*}, mark size={1.125 pt}, mark repeat={1}, mark options={color={rgb,1:red,0.0;green,0.0;blue,0.0}, draw opacity={1.0}, fill={rgb,1:red,0.192;green,0.51;blue,0.74}, fill opacity={1.0}, line width={0.75}, rotate={0}, solid}]
        table[row sep={\\}]
        {
            \\
            2.0  0.4438415692298939  \\
            3.0  0.5722665224339849  \\
            4.0  0.09741361346669566  \\
            5.0  0.3873011940216885  \\
            6.0  0.15199898233305448  \\
            7.0  0.0340535878418282  \\
            8.0  0.28271544207153304  \\
            9.0  0.12400199634914232  \\
        }
        ;
    \addplot[color={rgb,1:red,0.192;green,0.639;blue,0.329}, name path={57dc6f95-597f-4fd7-9386-891b1f7a5ff3}, draw opacity={1.0}, line width={1}, solid, mark={square*}, mark size={1.125 pt}, mark repeat={1}, mark options={color={rgb,1:red,0.0;green,0.0;blue,0.0}, draw opacity={1.0}, fill={rgb,1:red,0.192;green,0.639;blue,0.329}, fill opacity={1.0}, line width={0.75}, rotate={0}, solid}]
        table[row sep={\\}]
        {
            \\
            2.0  0.47655685471252596  \\
            3.0  0.3124148499808578  \\
            4.0  0.06456566644922809  \\
            5.0  0.3516049845676171  \\
            6.0  0.4876700638957033  \\
            7.0  0.47024683717300697  \\
            8.0  0.6575285480383295  \\
            9.0  0.01005193698282365  \\
        }
        ;
    \addplot[color={rgb,1:red,0.87;green,0.176;blue,0.149}, name path={49063129-3bae-4d60-b2c3-4213e0004547}, draw opacity={1.0}, line width={1}, dotted, mark={triangle*}, mark size={1.125 pt}, mark repeat={1}, mark options={color={rgb,1:red,0.0;green,0.0;blue,0.0}, draw opacity={1.0}, fill={rgb,1:red,0.87;green,0.176;blue,0.149}, fill opacity={1.0}, line width={0.75}, rotate={0}, solid}]
        table[row sep={\\}]
        {
            \\
            4.0  0.012476104406755091  \\
        }
        ;
    \addplot[color={rgb,1:red,0.87;green,0.176;blue,0.149}, name path={49063129-3bae-4d60-b2c3-4213e0004547}, draw opacity={1.0}, line width={1}, dotted, mark={triangle*}, mark size={1.125 pt}, mark repeat={1}, mark options={color={rgb,1:red,0.0;green,0.0;blue,0.0}, draw opacity={1.0}, fill={rgb,1:red,0.87;green,0.176;blue,0.149}, fill opacity={1.0}, line width={0.75}, rotate={0}, solid}, forget plot]
        table[row sep={\\}]
        {
            \\
            7.0  0.5835520250165347  \\
        }
        ;
    \addplot[color={rgb,1:red,0.87;green,0.176;blue,0.149}, name path={49063129-3bae-4d60-b2c3-4213e0004547}, draw opacity={1.0}, line width={1}, dotted, mark={triangle*}, mark size={1.125 pt}, mark repeat={1}, mark options={color={rgb,1:red,0.0;green,0.0;blue,0.0}, draw opacity={1.0}, fill={rgb,1:red,0.87;green,0.176;blue,0.149}, fill opacity={1.0}, line width={0.75}, rotate={0}, solid}, forget plot]
        table[row sep={\\}]
        {
            \\
            9.0  0.010030970758954514  \\
        }
        ;
\end{axis}
\end{tikzpicture}
        \caption{Step ratio to threshold.}
        \label{fig:adc_step}
    \end{subfigure}
    \caption{Evaluation metrics for the autonomous driving scenario with a comparable systems.}
    \label{fig:adc}
    \vskip -0.5 in
\end{figure*}

We solved each of the four safety validation tasks using 3 transfer learning algorithms and report the evaluation metrics against the task index in \cref{fig:gwl,fig:gwc,fig:adl,fig:adc}. The discussion of the results is grouped by evaluation metric.

\paragraph{Jumpstart.}
\Cref{fig:gwl_jumpstart,fig:gwc_jumpstart,fig:adl_jumpstart,fig:adc_jumpstart} show the jumpstart of the fine-tune and A2T architectures. Across all four safety validation problems, the transfer learning algorithms contributed a significant increase in initial performance. For most tasks, the A2T architecture had slightly better jumpstart than simply reusing the previous solution, especially in \cref{fig:gwc_jumpstart}. The gridworld with comparable systems had substantially different failure modes between tasks and therefore had the least benefit in jumpstart.  The safety validation problems with learning systems had the jumpstart decrease with the number of tasks observed, likely due to the increase in difficulty of the tasks. For the safety validation problems involving comparable systems, however, the jumpstart tended to increase with the number of source tasks. We hypothesize that with more source tasks, we are more likely to have a task that closely matches the current tasks, and can therefore immediately have reasonable performance. 

\paragraph{Final Performance.}
\Cref{fig:gwl_final,fig:gwc_final,fig:adl_final,fig:adc_final} show the final performance of each transfer learning algorithm. The final safety validation performance can be significantly improved by both A2T approaches, but not through fine-tuning. In all but \cref{fig:adl_final}, the fine-tuning approach was not able to match the no-transfer performance given the same number of iterations. The lack of performance could mean that the $Q$-network for one task is not learning a set of features that is useful for solving other tasks, so updating only the final layer does not provide enough capacity to solve the problem.  The A2T networks, however, are able to achieve significantly improved final performance, which generally increases with the number of source tasks. The A2T network with state and action value transformations outperforms the basic A2T network in both safety validation problems with comparable systems, which are the problems it was designed for. Both of the safety validation problems with a learning system show the maximal gain in final performance in the middle of the sequence of tasks. A lower gain in early tasks may be due to those tasks being easy to solve, while a lower gain in the later tasks may be due to only having a few failure modes to exploit. The middle tasks may be challenging to solve but may have a diversity of failure modes that the previous policies can help identify. More experimentation is needed to fully understand these trends. 

\paragraph{Steps to Threshold.}
\Cref{fig:gwl_step,fig:gwc_step,fig:adl_step,fig:adc_step} show the number of training steps required to reach the near-optimal performance of the no-transfer algorithm. In some cases (and especially for the fine-tune approach), near-optimal performance is never reached so those data points are omitted from the plots.  We observe that the number of training steps can be reduced by both A2T networks, but in different conditions. The basic A2T network performs well when validating a learning system because parts of previous solutions can be used directly. In \cref{fig:gwl_step}, the number of training steps on some tasks could be reduced by \num{50}\% and in \cref{fig:adl_step} the number of training steps is reduced by more than an order of magnitude in some cases.  

The A2T network with state and action transformations performs slightly worse than the basic A2T network in \cref{fig:gwl_step} and has similar performance in \cref{fig:adl_step}, but significantly outperforms the A2T network for many tasks in the comparable systems setting, which is the setting it was designed for. In \cref{fig:gwc_step}, the basic A2T network requires more steps than the no-transfer algorithm, which negates the utility of the more complex architecture, while the A2T+SAVT network was able to reduce the number of training steps by up to \num{50}\%. We note that generally, the fine-tune approach is unable to achieve the same performance as learning from scratch but when it does reach near-optimal performance, it requires fewer training steps than learning from scratch.

\paragraph{Summary.} From our experiments we conclude that transfer learning can be an effective strategy for improving performance and efficiency of safety validation algorithms. Transfer through fine-tuning can give a significant increase in jumpstart but often fails to reach the level of performance of a $Q$-network trained from scratch. The A2T networks also provides an increase in jumpstart as well as an increase in final performance. The use of a small attention network allows for quick adaptation to new domains as evidenced by the reduction in the number of training steps required to reach near-optimal performance. When the tasks differ significantly from each other, however, the basic A2T network may take longer than the no-transfer algorithm to reach near-optimal performance. We fix this problem by introducing state and action value transformations for each source solution and demonstrate improved training efficiency over the no-transfer algorithm.

\section{Conclusion}
The validation of safety-critical autonomous systems is crucial for their safe deployment. Existing algorithms for validation often start from scratch each time the system changes. The nature of system design implies that safety validation will be performed iteratively on related systems, and should therefore benefit from past experience. We formulate iterative safety validation as a transfer learning problem and demonstrate improvements in both efficiency and performance of transfer learning algorithms compared to a no-transfer baseline. We augmented the attend, adapt, and transfer algorithm with state and action value transformations to allow for more transfer between disparate tasks. We evaluated jumpstart, final performance, and steps to threshold metrics on four iterative safety validation problems in gridworld and autonomous driving domains. Future work will include exploring the failure modes discovered by each algorithm to gain insights into how transfer is occurring. These insights may help us understand under what conditions we can expect performance and efficiency improvements.

\bibliography{references.bib}

\end{document}